\documentclass[sigconf]{acmart}
\copyrightyear{2025}
\acmYear{2025}
\setcopyright{cc}
\setcctype{by-nc-sa}
\acmConference[KDD '25]{Proceedings of the 31st ACM SIGKDD Conference on Knowledge Discovery and Data Mining V.2}{August 3--7, 2025}{Toronto, ON, Canada}
\acmBooktitle{Proceedings of the 31st ACM SIGKDD Conference on Knowledge Discovery and Data Mining V.2 (KDD '25), August 3--7, 2025, Toronto, ON, Canada}
\acmDOI{10.1145/3711896.3736880}
\acmISBN{979-8-4007-1454-2/2025/08}

\usepackage{algorithm}
\usepackage{algorithmic}
\usepackage{url}            
\usepackage{booktabs}       
\usepackage{amsfonts}       
\usepackage{nicefrac}       
\usepackage{microtype}      
\usepackage{natbib}
\usepackage{bbding} 
\usepackage{graphicx}
\usepackage{adjustbox}
\usepackage{multicol}
\usepackage{multirow}
\usepackage{amsmath}
\usepackage{pifont} 
\usepackage{makecell} 
\usepackage{subcaption}
\usepackage{lineno}
\usepackage{colortbl}
\usepackage{thmtools}
\usepackage{enumitem}
\usepackage{bm}
\usepackage{theoremref}
\usepackage{amsthm}
\usepackage{hyperref}
\usepackage{balance}

\newcommand{\ie}[0]{\textit{i.e.},\ }   

\newcommand{\modelabbr}{CLID-MU}
\newcommand{\modelfull}{Cross-Layer Information Divergence Based Meta Update Strategy}
\newcommand\kddavailabilityurl{https://doi.org/10.5281/zenodo.15595961}

\theoremstyle{plain}

\newtheorem{theorem}{Theorem}
\newtheorem{remark}{Remark}

\theoremstyle{definition}
\newtheorem{definition}{Definition}

\AtBeginDocument{%
  }

\begin{document}

\title{CLID-MU: Cross-Layer Information Divergence Based Meta Update Strategy for Learning with Noisy Labels}

\author{Ruofan Hu}
\affiliation{%
  \institution{Worcester Polytechnic Institute}
  \city{Worcester}
  \state{MA}
  \country{USA}
}
\email{rhu@wpi.edu}

\author{Dongyu Zhang}
\affiliation{%
  \institution{ByteDance}
  \city{San Jose}
  \state{CA}
  \country{USA}
}
\email{dongyu.zhang@bytedance.com}

\author{Huayi Zhang}
\affiliation{
  \institution{ByteDance}\city{San Jose}\state{CA}\country{USA}
}
\email{huayi.zhang@bytedance.com}

\author{Elke Rundensteiner}
\affiliation{%
  \institution{Worcester Polytechnic Institute}
  \city{Worcester}
  \state{MA}
  \country{USA}
}
\email{rundenst@wpi.edu}

\begin{abstract}
Learning with noisy labels (LNL) is essential for training deep neural networks with imperfect data. Meta-learning approaches have achieved success by using a clean unbiased labeled set to train a robust model. However, this approach heavily depends on the availability of a clean labeled meta-dataset, which is difficult to obtain in practice. In this work, we thus tackle the challenge of meta-learning for noisy label scenarios without relying on a clean labeled dataset. Our approach leverages the data itself while bypassing the need for labels. Building on the insight that clean samples effectively preserve the consistency of related data structures across the last hidden and the final layer, whereas noisy samples disrupt this consistency, we design the Cross-layer Information Divergence-based Meta Update Strategy (CLID-MU). CLID-MU leverages the alignment of data structures across these diverse feature spaces to evaluate model performance and use this alignment to guide training. Experiments on benchmark datasets with varying amounts of labels under both synthetic and real-world noise demonstrate that CLID-MU outperforms state-of-the-art methods. The code is released at \href{https://github.com/ruofanhu/CLID-MU}{\textit{https://github.com/ruofanhu/CLID-MU}}.
\end{abstract}




\maketitle

\ifdefempty{\kddavailabilityurl}{}{
\begingroup\small\noindent\raggedright\textbf{KDD Availability Link:}\\
The source code of this paper has been made publicly available at \url{\kddavailabilityurl}.
\endgroup
}

\section{Introduction}
\textbf{Background.}
Developing deep neural networks (DNN) requires a large amount of labeled data. Yet due to the high cost and difficulty of data labeling, curating large datasets with high-quality labels is challenging. Some real-world datasets are curated via web crawling or crowdsourcing, inevitably yielding noisy labels \cite{song2022learning}. Recent advances in learning with noisy labels (LNL) have shown success in training robust models under such conditions \cite{song2022learning}. Among them, meta-learning approaches \cite{Zhang2020SemiSupervisedLW,Wang2022ImbalancedSL} are particularly effective by leveraging \textit{a small clean unbiased (balanced) labeled set} as a meta-dataset to evaluate model performance during training and effectively guide the training process.

\textbf{State-of-the-Art and its limitations.} 
However, acquiring a high-quality, unbiased labeled dataset is often infeasible in real-world scenarios due to the substantial time, cost, and effort required. For example, the popular CIFAR-100 dataset consists of images categorized into 100 classes, including various animal species and vehicle types. Selecting an equal number of images per class and accurately assigning labels would be highly time-consuming and labor-intensive. Moreover, meta-learning-based methods have been shown to perform poorly when using a noisy labeled meta-dataset instead of a clean one \cite{zhu-etal-2023-weaker}, posing a major obstacle in real-world applications where label noise is unavoidable. Existing studies have explored two main strategies: (1) using a noisy labeled set with a robust loss function, such as MAE \cite{ghosh2021we,yong2022holistic}, as meta-loss to mitigate the impact of noisy labels, and (2) progressively selecting a pseudo-clean subset as the meta-dataset during training utilizing the small-loss trick \cite{Zhang2021LearningFS,sun2024variational,xu2021faster}. However, the effectiveness of these methods is significantly hindered by the noisy patterns of the datasets. Robust loss functions struggle to handle complex noise patterns, such as instance-dependent noise \cite{wang2021learning}. While pseudo-clean subset selection methods often require careful threshold tuning, the small-loss trick fails to reliably distinguish between clean and noisy labeled samples under instance-dependent noise. Consequently, the evaluation of model performance based on robust loss functions or pseudo-clean subsets becomes unreliable. The guidance derived from such unreliable evaluation approaches can mislead the training process, increasing the risk of overfitting to noisy labels.

\textbf{Challenges.} The primary challenge arises from the noisy labels. In the absence of clean labels, it becomes impractical to establish well-founded criteria for selecting pseudo-clean subsets. Prior works typically rely on conventional supervised loss functions as the meta-objective to update the meta-model. However, the meta-update process is susceptible to the potential noisy labels in the meta-dataset, which causes bias propagation to the meta-model. Consequently, a solution that operates independently of noisy labels must be developed.

\textbf{Proposed method.} 
In this paper, we first propose an unsupervised metric called Cross-Layer Information Divergence (CLID), which operates independently of labels. CLID leverages the insight that clean samples maintain alignment between the data distribution in the penultimate latent space and the output layer, whereas noisy samples tend to disrupt this alignment. Our CLID metric measures the divergence of the data distribution at the last hidden layer and the final layer of the model. We demonstrate that CLID indeed closely correlates with the model performance. 
Building on this, we introduce a novel CLID-based meta-update strategy, termed CLID-MU, that addresses the above challenge of meta-learning with noisy labels in the absence of clean labeled data. 
CLID-MU exploits the alignment of data structures across diverse feature spaces and is designed to function independently of label quality, ensuring robust model performance and producing more compact features. 
The core idea is to dynamically measure the CLID of the model for each training batch, providing informative guidance to the model training process. Specifically, the meta-model is updated using the meta-gradients derived from CLID calculations on the data itself (independent of the labels). The meta-model, in turn, offers valuable signals to enhance the performance of the classification model.

\textbf{Contributions.} Our contributions include the following: 
\begin{itemize}[leftmargin=*, itemsep=2pt, topsep=2pt]
    \item We propose Cross-Layer Information Divergence (CLID), a novel unsupervised evaluation metric designed for scenarios lacking clean labeled data.
    \item We introduce CLID-MU, a CLID-guided meta-update strategy for meta-learning with noisy labels, without requiring clean validation data.
    \item Extensive experiments on benchmark datasets with synthetic and real-world noise show that CLID-MU consistently outperforms state-of-the-art methods.
\end{itemize}

\section{Related Work}


Numerous methods have been proposed to train robust deep networks with noisy labels. Easy-to-plug-in solutions like robust-loss functions,  MAE \cite{Ghosh2017RobustLF}, GCE \cite{Zhang2018GeneralizedCE}, and APL \cite{Ma2020NormalizedLF}, aim to resist label noise, but they still overfit when noise levels are high or complex. Similarly, regularization terms \cite{liu2020early, Yi2022OnLC,zhang2021lancet} are added to the loss function to reduce overfitting implicitly. Loss correction methods adjust sample loss based on noise transition matrices during training \cite{xia2019anchor,yao2020dual,yong2022holistic}, while other strategies reduce weights for noisy samples \cite{jiang2018mentornet,ren2018learning,hofmann2022demonstration}. Hybrid methods like CoLafier \cite{zhang2024colafier}, DISC \cite{Li2023DISCLF}, and UNITY \cite{
hofmann2025agree} incorporate both clean sample selection and label correction. However, these methods involve complex training procedures, requiring the coordination of multiple models and the careful tuning of dataset-specific hyperparameters, which makes them difficult to apply in practice.

Meta-learning \cite{ren2018learning, Shu2019MetaWeightNetLA,hu2023uce,zhang2021elite,zhang2024metastore,wu2024rose} is a general approach for learning with imperfect data. These methods optimize various configurations by using a clean validation set to evaluate the model, such as the weight for each training sample \cite{ren2018learning}, the label transition matrix \cite{yao2020dual}, the explicit weighting function \cite{Shu2019MetaWeightNetLA,sun2024variational} for example re-weighting, the teacher model parameter \cite{taraday2023enhanced} for label correction. Due to limited resources, constructing a clean and balanced validation set using expert knowledge is often impractical. To eliminate the need for a clean validation set, recent approaches employ robust loss functions on noisy labels \cite{ghosh2021we,yong2022holistic} or utilize heuristic approaches to select presumably clean samples as a validation set \cite{Zhang2021LearningFS,sun2024variational}. Despite their promise, these methods encounter a performance ceiling when handling complex noise patterns, primarily due to their reliance on the quality of the labels. This dependency can result in overfitting to noise and hinder generalization. This highlights the need for more robust meta-learning approaches that can effectively deal with this challenging yet realistic problem.

Model selection without a clean validation set is a known challenge in weakly supervised settings like semi-supervised learning (SSL) and partial-label learning (PLL). Recent methods attempt to address this using validation-free strategies. For example, SLAM \cite{li2024towards} and QLDS \cite{feofanov2023random} estimate generalization errors in SSL. PLENCH \cite{wang2025realistic} benchmarks PLL methods and proposes new selection criteria. However, these methods, being non-differentiable, are not suitable for gradient-based meta-learning.

\section{Preliminaries}

\subsection{Problem Formulation} \label{p_def}
Let $D=\{x_{i}\}_{i=1}^N$ denote an unlabeled dataset drawn from a distribution $(x_i,y_i) \sim \mathcal{X}\times\mathcal{Y}$, where $y_i \in \{0,1\}^c$ is the one-hot ground truth label of $x_i$ over $c$ classes. With weak labelers such as crowdsourced workers, $D$ is converted to a {\it noisy training set}  $\Tilde{D}=\{(x_{i},\Tilde{y}_{i})\}_{i=1}^{N}$. $\Tilde{D}$ may contain \textit{inaccurate labels}, where $\Tilde{y}_{i} \neq y_i$.
We assume no clean labeled subset (\ie validation set) is available in $\Tilde{D}$.

Given a noisy labeled dataset $\Tilde{D}$, our goal is to develop a 
classification model that, without access to clean labeled data, can correctly predict the labels of unseen test data. The classification model $f_{\theta}$ is formulated as $f_{\theta}(x) = f_{\theta_{2}}^{cls}\circ f_{\theta_{1}}^{ext}(x)$ on instance $x$, where $f^{ext}_{\theta_{1}}$ is a feature extractor and $f^{cls}_{\theta_{2}}$ is a classifier. 
Let $\boldsymbol{z}=f^{ext}_{\theta_{1}}(x)$ denote the feature embedding of $x$ and $\boldsymbol{q}=f^{cls}_{\theta_{2}}(\boldsymbol{z})$  the class probability, with $\boldsymbol{z}$ and $\boldsymbol{q}$ residing in the output space of the last hidden layer and final (output) layer, respectively. 

 
\subsection{Meta-learning Procedure}



Here, we introduce the preliminaries on
meta-learning upon
which our proposed method rests \cite{Shu2019MetaWeightNetLA,ghosh2021we,yong2022holistic,ren2018learning}. In meta-learning for noisy labels, there is a noisy training set $\Tilde{D}=\{(x_{i},\Tilde{y}_{i})\}_{i=1}^{N}$, and a separate meta-dataset $D^{meta} = \{(x_j^{meta},y_j^{meta})\}_{j=1}^M$, where $M \ll N$ and $y_j^{meta}$ could be inaccurate \cite{Tu2023LearningFN,ghosh2021we}. 
Below, we explain the main strategy of reweighting training samples using
an example based on the procedure outlined in WNet \cite{Shu2019MetaWeightNetLA}.

Namely, a meta-model $\Omega(\cdot;\psi)$ is coupled with the classification model $f(\cdot;\theta)$ to learn a weight for each training example. The meta-model, instantiated as a multilayer perceptron network (MLP), takes the training loss as input and maps it to a weight for the training sample. This mapping allows the meta-model to dynamically adjust the importance of each training sample during the training process.
The parameter $\theta^*$ is optimized by minimizing the weighted loss:
\begin{equation} \label{eq:lt_op}
\theta^*(\psi) = \arg \min_{\theta} \frac{1}{N} \sum_{i=1}^{N}\Omega(L_{i}(\theta);\psi) \cdot L_{i}(\theta),
\end{equation} 
where $L_i(\theta)=l(f(x_{i};\theta);\Tilde{y}_{i})$ denotes cross-entropy loss for the $i$-th training sample and $\Omega(L_{i}(\theta);\psi)$ represents the corresponding generated weight for that sample.

\textbf{Meta objective.} Eq.(\ref{eq:lv_sup}) denotes the meta loss, where $l^{meta}$ could be the cross-entropy loss (CE) or mean absolute error (MAE). The optimal parameter $\psi^*$ for the meta-model can be obtained by minimizing the meta loss defined in Eq.(\ref{eq:lv_sup_op}).

\begin{equation} \label{eq:lv_sup}
L^{meta}_j(\theta^*(\psi)) =l^{meta}(f(x^{meta}_{j};\theta^*(\psi));y^{meta}_{j}).
\end{equation}
\begin{equation} \label{eq:lv_sup_op}
\psi^* = \arg \min_{\psi} \frac{1}{M}\sum_{i=j}^{M}L^{meta}_{j}(\theta^*(\psi)).
\end{equation}

\textbf{Bi-level optimization.} To solve both Eq.(\ref{eq:lt_op}) and Eq.(\ref{eq:lv_sup_op}), an online updating strategy is widely used in the meta-learning literature \cite{zhang2024introduction}
to update $\psi$ and $\theta$ iteratively. Consider the $t$-th iteration, three steps are involved: {\it Virtual-Train, Meta-Train},  and {\it Actual-Train}. 
First, a batch of labeled samples $\{(x_{i},\Tilde{y}_{i})\}_{i=1}^{n}$ 
and  meta-dataset $ \{(x_{j}^{meta},y_{j}^{meta})\}_{j=1}^m$
are sampled, $n$ and $m$ represent the batch sizes, respectively.  We may approximate $\theta^*$ and $\psi^*$ with one gradient descent step updated value via a first-order Taylor expansion of the loss function. 
In the {\it Virtual-Train} step, the update of classification model's parameter $\theta$ is formulated as:
\begin{equation} \label{eq:virtaul-train}
\hat{\theta}^{t+1}(\psi)=\theta^{t}-\alpha \frac{1}{n}\sum_{i=1}^{n}\Omega(L_{i};\psi^t)\nabla_{\theta} L_{i}(\theta)\mid_{\theta^{t}},
\end{equation}
where $\alpha$ is the learning rate for the classification model. Then  {\it Meta-Train} updates the meta-model  by:
\begin{equation} \label{eq:meta-train}
\psi^{t+1} =\psi^{t} -\gamma \frac{1}{m}\sum_{j=1}^{m}\nabla_{\psi} L_{j}^{meta}(\hat{\theta}^{t+1}(\psi))\mid_{\psi^t},
\end{equation}
where $\gamma$ is the learning rate for the meta-model. In the 
{\it Actual-Train} step, the classification model is  finally updated using the updated meta-model by:
\begin{equation} \label{eq:actual-train}
\theta^{t+1}=\theta^{t}-\alpha \frac{1}{n}\sum_{i=1}^{n}\Omega(L_{i};\psi^{t+1})\nabla_{\theta} L_{i}(\theta)\mid_{\theta^{t}},
\end{equation}

\begin{figure}[htbp]
  \centering

    \begin{subfigure}[b]{0.49\columnwidth}
    \centering
    \includegraphics[width=\linewidth]{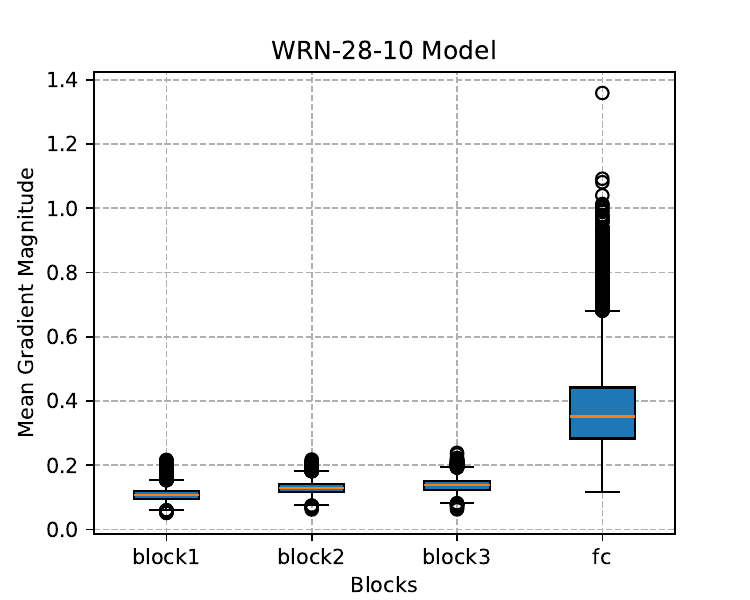}
    \Description{ The gradient magnitude of the classifier layer is larger than that of the hidden layer in WRN-28-10 and Resnet-32 models.}
    \label{fig:mg_wrn2810}
    \end{subfigure}
  \hspace{0cm} 
    \begin{subfigure}[b]{0.49\columnwidth}
    \centering
    \includegraphics[width=\linewidth]{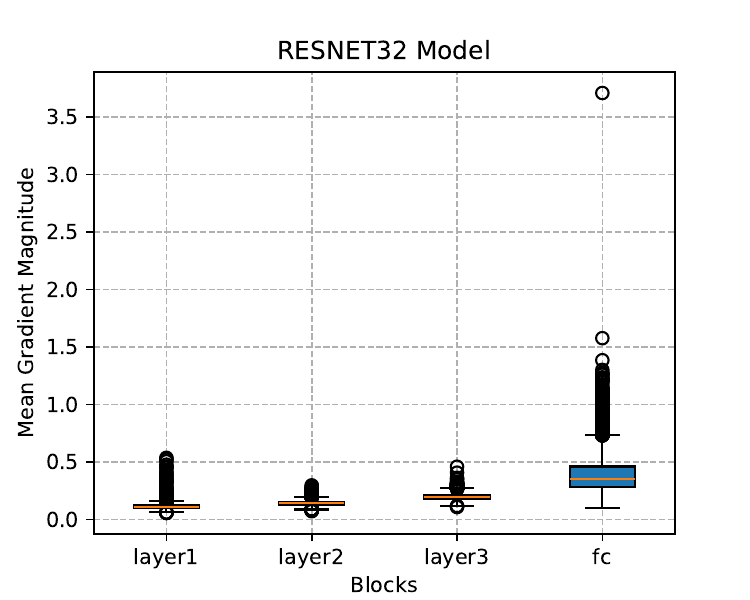}
    \label{fig:mg_res32}
    \end{subfigure}
  
  \caption{Illustration of gradient magnitudes in WRN-28-10 and ResNet32 residual layer blocks.}
  \label{fig: mgs}
\end{figure}
\section{Our Proposed Method: CLID-MU}
In this section, we present our proposed method \modelabbr{}. Our method builds on the  \textit{cluster assumption}, namely, samples forming a structure are more likely to belong to the same class \cite{Zhou2003LearningWL}.  
We postulate that clean training samples align the data's structure in the feature space with that in the label space. To assess this alignment, we propose an unsupervised metric, CLID, which measures the divergence between the data distributions in the feature space and the label space. 
We then demonstrate how our proposed CLID metric correlates with the classification performance of DNNs. This important insight allows us to utilize our proposed differentiable metric effectively for non-supervised meta-learning. 

\subsection{Cross-Layer Information Divergence (CLID)} \label{sec:clid}
Given a batch of data $\{x_j\}_{j=1}^m$ and a classification model $f_{\theta}$, we create the feature embedding $\{\boldsymbol{z}_j\}_{j=1}^m$ and class probability $\{\boldsymbol{q}_j\}_{j=1}^m$, which are data representations generated from the last hidden layer and final layer of DNN.

We generate a {\it fully connected embedding graph} $\boldsymbol{G}^e$ to capture the similarity of samples in the latent space as:
\begin{equation} \label{eq:G_embedding}
G_{ij}^e = \text{exp}(\text{cos}(\boldsymbol{z}_i ,\boldsymbol{z}_j)/\tau), \quad \forall i,j \in \{1, \ldots, m\},
\end{equation}
where $\tau$ is a hyperparameter for temperature scaling and the exponential function is used to emphasize strong similarities.
We then build the {\it class probability graph} by constructing  the similarity matrix $\boldsymbol{G}^q$ as:

\begin{equation} \label{eq:G_class}
G_{ij}^q = \text{cos}(\boldsymbol{q}_i,\boldsymbol{q}_j) \quad \forall i,j \in \{1, \ldots, m\}
\end{equation}

Recall that we aim to measure the divergence of the representations produced by different DNN layers, while each graph represents instead the similarities between the representations. To better model the global structure of the two graphs and represent a valid probability distribution, we normalize 
both $\boldsymbol{G}^e$ and $\boldsymbol{G}^q$ with $\hat{G}_{ij}:= G_{ij}/\sum_j{G_{ij}}$.

Given that we have constructed two graphs, each representing a data distribution, we can now measure the cross-layer information divergence between the two normalized graphs. Since the evolution speed (gradient magnitude) of each layer differs, layers with larger gradient magnitudes learn more information during each training step. Consequently, the data distribution generated by the slower-updating layer should gradually align with that of the faster-updating layer. We find that the gradient magnitude of the classifier layer is larger than that of the hidden layer, as shown in Figure \ref{fig: mgs}. Therefore, we compute CLID using the cross-entropy between the two normalized graphs, as shown below:
\begin{equation} \label{eq:clid_loss}
L^{clid}= \frac{1}{m^2}\sum_{i=1}^m\sum_{j=1}^m{-\hat{G}_{ij}^q\text{log}\hat{G}_{ij}^e}
\end{equation}

\begin{figure*}[htbp]
    \centering
    \begin{subfigure}{0.33\textwidth}
        \centering
        \includegraphics[width=\textwidth]{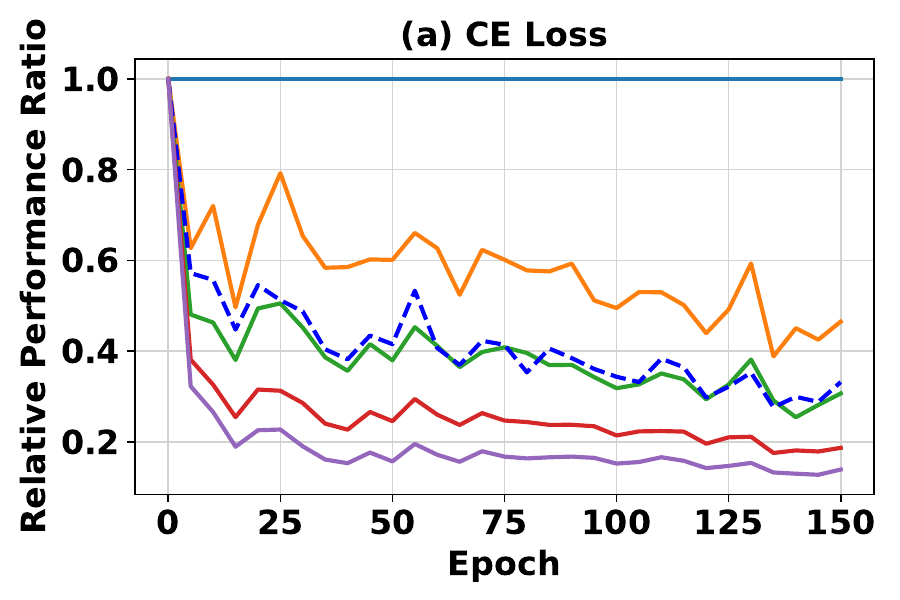} 
        \label{fig:sub1}
    \end{subfigure}
    \hfill
    \begin{subfigure}{0.33\textwidth}
        \centering
        \includegraphics[width=\textwidth]{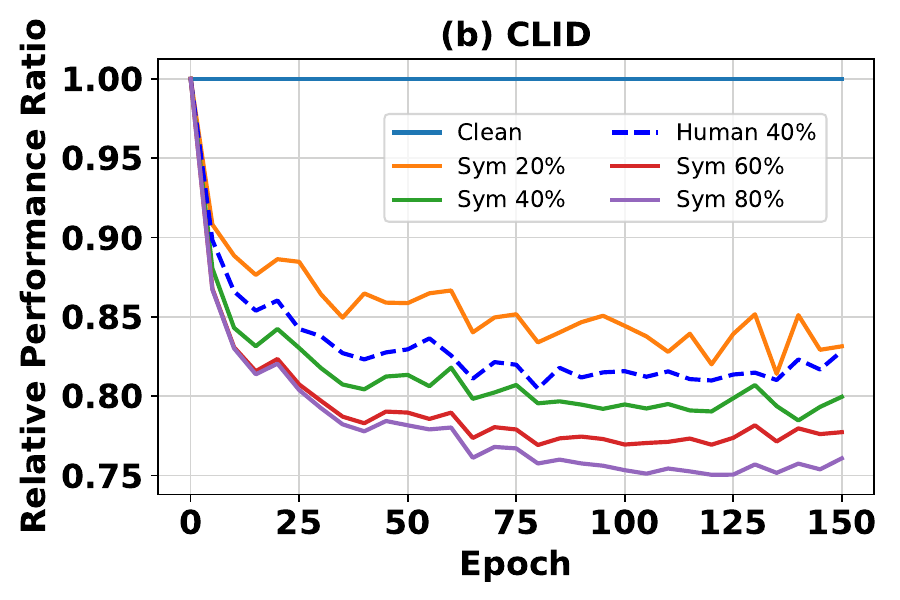} 
        \label{fig:sub2}
    \end{subfigure}
    \hfill
    \begin{subfigure}{0.33\textwidth}
        \centering
        \includegraphics[width=\textwidth]{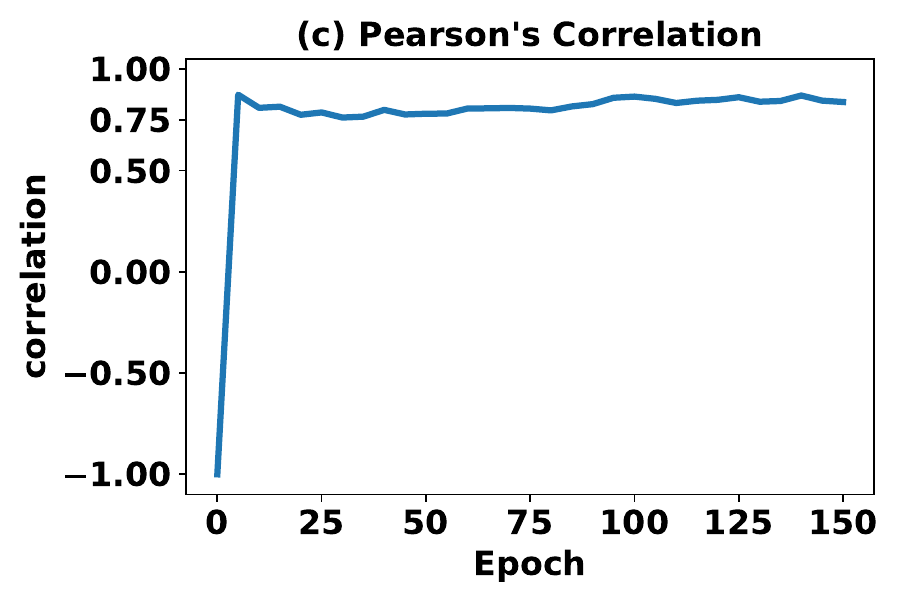} 
        \label{fig:sub3}
    \end{subfigure}
    \Description{The Relative Performance Ratio (RPR) of CE loss and
CLID on the test set exhibits similar trends under various noise settings on the CIFAR-10 test set.}
    \caption{Demonstration on CIFAR-10: The relative performance ratio of (a) Cross-Entropy (CE) loss, (b) CLID, and (c) Pearson's correlation between CE loss and CLID across all data settings at each epoch.}
    \label{fig:ce_clid_cifar10}
\end{figure*}

\begin{figure*}[htbp]
    \centering
    \begin{subfigure}{0.33\textwidth}
        \centering
        \includegraphics[width=\textwidth]{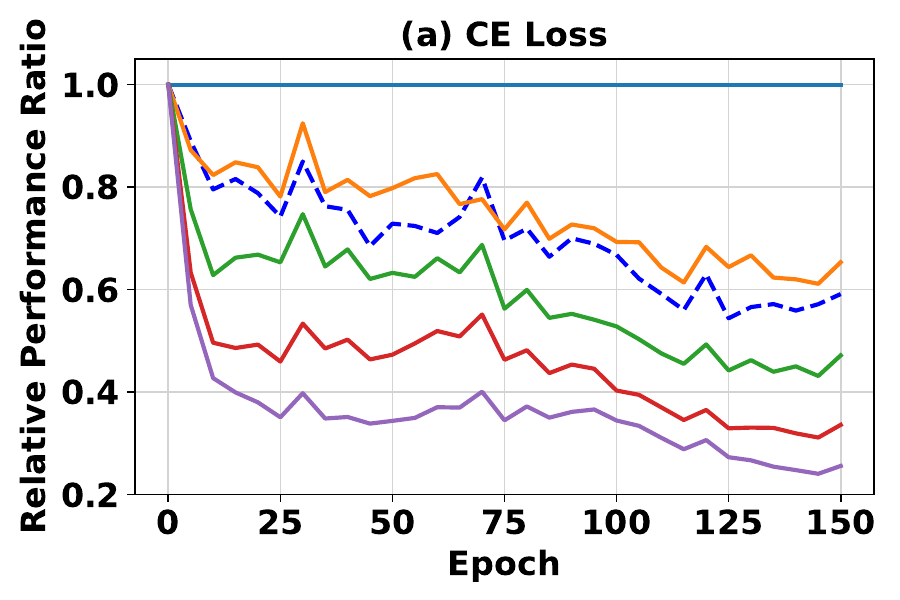} 
        \label{fig:sub1_}
    \end{subfigure}
    \hfill
    \begin{subfigure}{0.33\textwidth}
        \centering
        \includegraphics[width=\textwidth]{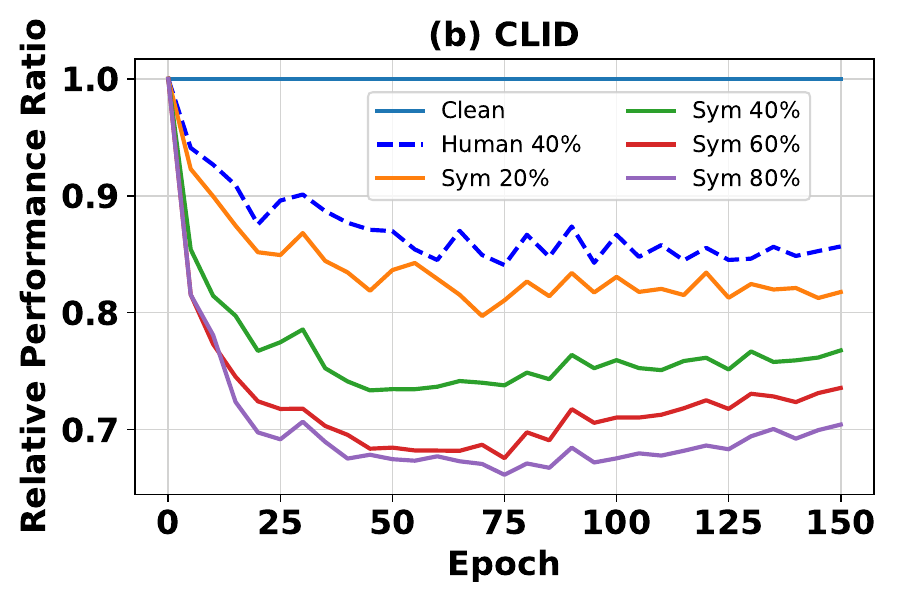} 
        \label{fig:sub2_}
    \end{subfigure}
    \hfill
    \begin{subfigure}{0.33\textwidth}
        \centering
        \includegraphics[width=\textwidth]{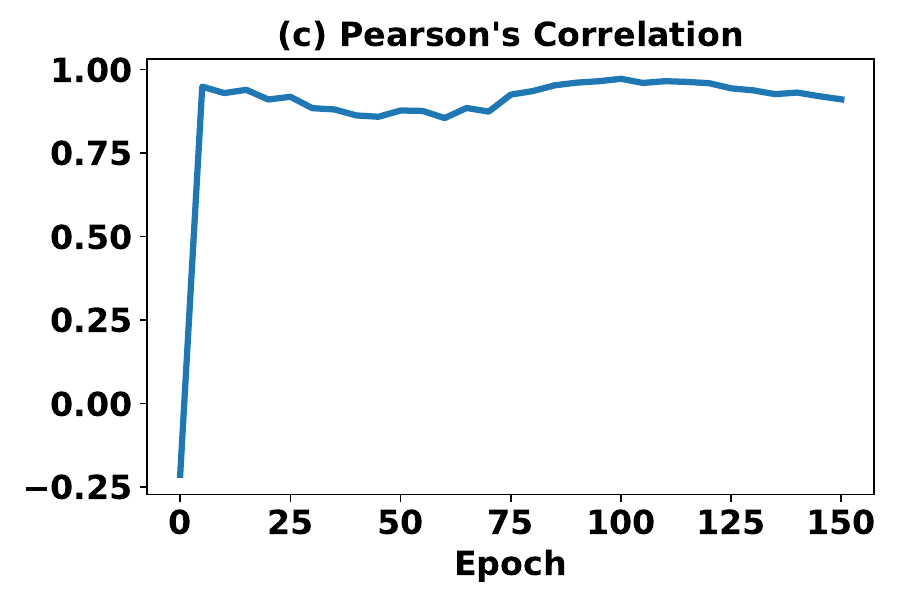} 
        \label{fig:sub3_}
    \end{subfigure}
    \Description{The Relative Performance Ratio (RPR) of CE loss and
CLID on the test set exhibits similar trends under various noise settings on the CIFAR-100 test set.}
    \caption{Demonstration on CIFAR-100: The relative performance ratio of (a) Cross-Entropy (CE) loss, (b) CLID, and (c) Pearson's correlation between CE loss and CLID across all data settings at each epoch.}
    \label{fig:ce_clid_cifar100}
\end{figure*}

\subsection{CLID and Model Performance} \label{sec:clid_performance}
The cross-entropy loss on the clean labeled test set is usually used to evaluate model performance. To establish a relationship between our novel unsupervised CLID metric and this standard supervised cross-entropy loss, we define two empirical alignment measures: the Relative Performance Ratio (RPR) and Performance Pearson's Correlation, as defined below.

\begin{definition}[Relative Performance Ratio (RPR)]
Let $\mathcal{D}$ denote a dataset, and let $f_\theta^n$ and $f_\theta^{\text{clean}}$ represent models trained on the dataset under noisy setting $n$ and perfect clean labels, respectively. Denote the test performance of $f_\theta^n$ as $P^n$ and the test performance of $f_\theta^{\text{clean}}$ as $P^{\text{clean}}$.
The \emph{Relative Performance Ratio (RPR)} of the model trained on setting $n$ is defined as: $\text{RPR}(n) = \frac{P^{{\text{clean}}}}{P^n}$.

\end{definition}

The RPR quantifies the relative performance degradation of a model trained under a noisy or altered setting compared to the ideal clean label scenario.
\begin{definition}[Performance Pearson's Correlation]
Let $\mathcal{D}$ be a dataset, and $\mathcal{N}$ be a set of noisy settings applied to $\mathcal{D}$. For a deep neural network $f_\theta$ trained on $(\mathcal{D}, \mathcal{N})$ under a fixed training protocol, let $A_t(\mathcal{N})$ and $B_t(\mathcal{N})$ denote two evaluation metrics measured at epoch $t$ for each noisy setting $n \in \mathcal{N}$.
We define the correlation between $A$ and $B$ at training epoch $t$ as the Pearson correlation coefficient:
\[
r(A_t(\mathcal{N}), B_t(\mathcal{N})) = \frac{\sum_{n} (A_t(n) - \overline{A_t})(B_t(n) - \overline{B_t})}{\sqrt{\sum_{n} (A_t(n) - \overline{A_t})^2} \sqrt{\sum_{n} (B_t(n) - \overline{B_t})^2}},
\]
where $\overline{A_t}$ and $\overline{B_t}$ are the mean values of $A_t(\mathcal{N})$ and $B_t(\mathcal{N})$ across all $n \in \mathcal{N}$, respectively. For all training epoch $t \in \{1, 2, \dots, T\}$, we say that $A$ and $B$ exhibit a strong correlation if $r(A_t(\mathcal{N}), B_t(\mathcal{N})) \geq \rho$,
for some threshold $\rho \in [0, 1]$, where $\rho$ represents a strong positive correlation (e.g., $\rho > 0.7$).
\end{definition}

\begin{figure*}[htbp]
    \centering
    \includegraphics[width=0.8\textwidth]{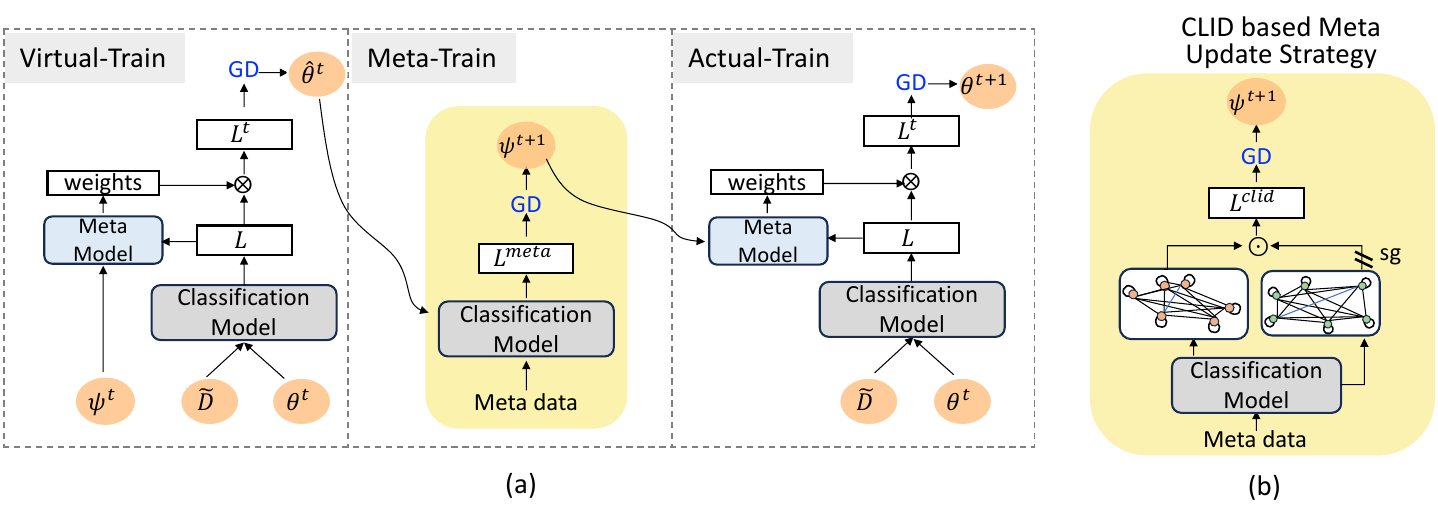}
    \Description[method flow]{(a) Illustration of the three major steps of a meta-learning method, using a reweight-based approach as an example. (b) We propose a new meta loss, CLID, for the Meta-Train step. Given an unlabeled dataset, a class probability graph and an embedding graph are constructed to measure the similarity between samples in their respective spaces. CLID measures the divergence of the data distribution of the two graphs. The sg denotes stop-gradient.}
    \caption{(a) Illustration of the three major steps of a meta-learning method, using a reweight-based approach as an example. (b) We propose a new meta loss, CLID, for the Meta-Train step. Given an unlabeled dataset, a class probability graph and an embedding graph are constructed to measure the similarity between samples in their respective spaces. CLID measures the divergence of the data distribution of the two graphs. The sg denotes stop-gradient.}
    \label{fig:method_flow}
\end{figure*}

We empirically demonstrate that CLID correlates with model performance on real-world datasets. Specifically, we explore a Resnet-34 \cite{He2015DeepRL} model on the CIFAR-10 and CIFAR-100 \cite{krizhevsky2009learning} datasets using 50,000 labeled samples and then evaluate the model using 10,000 test samples. The model's performance is assessed under various noise conditions, including noisy labels generated with symmetric noise ratios of \{0, 20\%, 40\%, 60\%, 80\%\}, where the correct label is randomly replaced with one of the other classes. Additionally, we consider noisy labels introduced by human annotators, with 40.21\% and 40.20\% of labeled samples affected (CIFAR-10N Worst and CIFAR-100N Fine \cite{wei2021learning}).

We compute the Relative Performance Ratio (RPR) of CE loss and CLID on the test set. Notably, the computation of CLID does not rely on clean labels. Thus, its value remains the same whether evaluated on clean or noisy data. The RPR of CE loss and CLID, as shown in Figures \ref{fig:ce_clid_cifar10}(a)(b) and Figures \ref{fig:ce_clid_cifar100}(a)(b), exhibit similar trends. This indicates that CLID effectively captures the performance degradation of models trained under various noisy settings, offering insights from the perspective of data structure alignment across different feature spaces. Further, the Performance Pearson's Correlations depicted in Figures \ref{fig:ce_clid_cifar10}(c) and \ref{fig:ce_clid_cifar100}(c) demonstrate a strong correlation between model performances measured by CLID and CE loss throughout the training process. This consistency underscores the agreement between the two metrics in evaluating model performance. Therefore, we can conclude that CLID is a robust and effective metric for assessing model performance.

 

\definecolor{lightgray}{gray}{0.95}

\begin{table*}[ht]
    \centering
    \caption{Comparison across meta-learning methods.}
    \begin{subtable}{\linewidth}
        \caption{Test accuracy (mean and std dev over 3 data folds) on CIFAR-10 and CIFAR-100 using WNet variants under symmetric (Sym.), asymmetric (Asy.), and instance-dependent noise (IDN) conditions. Only WNet-CE uses a clean meta-dataset. Bolded values indicate the highest and those within one standard deviation of the highest in each column.}
        \label{tab:wnet}
\resizebox{\textwidth}{!}{
\begin{tabular}{cc|ccccc|ccccc} \toprule
                                                    &                                  & \multicolumn{5}{c|}{CIFAR-10}                                                                                                                          & \multicolumn{5}{c}{CIFAR-100}                                                                                                                         \\
\multirow{-2}{*}{Method}                            &                                  & \multicolumn{1}{c}{Sym. 20\%} & \multicolumn{1}{c}{Sym. 40\%} & \multicolumn{1}{c}{Sym. 60\%} & \multicolumn{1}{c}{Asy. 40\%} & \multicolumn{1}{c|}{IDN 40\%} & \multicolumn{1}{c}{Sym. 20\%} & \multicolumn{1}{c}{Sym. 40\%} & \multicolumn{1}{c}{Sym. 60\%} & \multicolumn{1}{c}{Asy. 40\%} & \multicolumn{1}{c}{IDN 40\%} \\ \midrule
                                                    & best                             & \bm{$94.00_{\pm.24}$}              & $91.77_{\pm.14}$              & $86.68_{\pm.06}$              & $91.54_{\pm.02}$               & $91.19_{\pm.52}$               & $77.42_{\pm.07}$              & $73.11_{\pm.07}$              & $64.87_{\pm.32}$              & $62.49_{\pm.31}$               & $71.24_{\pm.52}$               \\
\multirow{-2}{*}{WNet-CE}                              & ens                         & \bm{$94.44_{\pm.21}$}              & \bm{$92.17_{\pm.25}$}              & $87.11_{\pm.18}$              & $91.98_{\pm.03}$               & $91.40_{\pm.62}$                & \bm{$78.46_{\pm.06}$}              & \bm{$74.00_{\pm.24}$}               & \bm{$66.00_{\pm.50}$}                & $64.73_{\pm.07}$               & \bm{$72.87_{\pm.66}$}               \\ \hline

                                                    & best                             & $94.92_{\pm.07}$              & $92.61_{\pm.27}$              & $88.38_{\pm.08}$              & $92.08_{\pm.16}$               & $90.53_{\pm.75}$               & $75.82_{\pm.73} $              & $70.30_{\pm1.1}$              & $63.06_{\pm.73}$              & $61.85_{\pm.63}$               & $68.16_{\pm.84}$               \\
\multirow{-2}{*}{WNet-CE(n)}                              & ens                         & $67.10_{\pm.91}$              & $56.47_{\pm.06}$              & $46.09_{\pm1.09}$              & $51.72_{\pm2.4}$               & $53.32 _{\pm.60}$                & $26.50_{\pm1.81}$              & $14.54_{\pm1.46}$               & $9.39_{\pm0.59}$                & $16.09_{\pm1.32}$               & $16.69_{\pm.94}$               \\ \hline

                                                    & best                             & $94.02_{\pm.03}$              & $91.70 _{\pm.21}$               & $86.80 _{\pm.33}$               & $91.90 _{\pm.04}$                & $91.30 _{\pm.78}$                & $76.98_{\pm.22}$              & $72.32_{\pm.18}$              & $63.60 _{\pm71}$               & $62.70 _{\pm.17}$                & $70.30 _{\pm.23}$                \\ 
\multirow{-2}{*}{WNet-CE(p)}                          & ens                         & \bm{$94.42_{\pm.01}$}              & \bm{$92.22_{\pm.44}$}              & \bm{$87.52_{\pm.11}$}              & \bm{$92.02_{\pm.07}$}               & $91.54_{\pm.53}$               & $78.05_{\pm.07}$              & $73.57_{\pm.04}$              & $64.34_{\pm.42}$              & $64.12_{\pm.52}$               & $71.54_{\pm.24}$               \\ \hline
                                                    & best                             & $94.10_{\pm.17}$               & $91.66_{\pm.35}$              & $86.88_{\pm.13}$              & $91.65_{\pm.18}$               & $91.42_{\pm.46}$               & $77.16_{\pm.16}$              & $72.63_{\pm.08}$              & $63.51_{\pm.31}$              & $63.74_{\pm.10}$                & $70.90 _{\pm.08}$                \\
\multirow{-2}{*}{WNet-MAE}                          & ens                         & \bm{$94.31_{\pm.07}$}              & \bm{$92.09_{\pm.42}$}              & $87.29_{\pm.08}$              & \bm{$92.12_{\pm.07}$}               & $91.76_{\pm.35}$               & \bm{$78.20_{\pm.26}$}               & $73.72_{\pm.04}$              & $64.41_{\pm.34}$              & $65.08_{\pm.28}$               & \bm{$72.72_{\pm.66}$}               \\ \hline 
\rowcolor{lightgray}                           & best     & $94.18_{\pm.23}$              & $91.66_{\pm.01}$              & $86.88_{\pm.01}$             & $91.97_{\pm.04}$               & $91.83_{\pm.21}$               & $77.36_{\pm.02}$              & $73.12_{\pm.38}$              & $64.82_{\pm.43}$              & $65.78_{\pm.1}$                & $71.50_{\pm.40}$                 \\ 
\rowcolor{lightgray} \multirow{-2}{*}{WNet-CLID} & ens & \bm{$94.34_{\pm.06}$}              & \bm{$92.22_{\pm.03}$}              & \bm{$87.27_{\pm.17}$}              & $91.93_{\pm.06}$               & \bm{$92.28_{\pm.11}$}               & \bm{$78.40 _{\pm.01}$}               & \bm{$74.32_{\pm.51}$}              & \bm{$65.98_{\pm.54}$}              & \bm{$67.66_{\pm.37}$}               & \bm{ $73.10_{\pm.32}$}   \\            
\bottomrule 
\end{tabular}
}

    \end{subtable}

    \vspace{1em} 

    \begin{subtable}{\linewidth}
    \centering
        \caption{Test accuracy (mean and std dev over 3 data folds) on CIFAR-10 and CIFAR-100 using VRI variants under symmetric (Sym.), asymmetric (Asy.), and instance-dependent noise (IDN) conditions. Only VRI-CE uses a clean meta-dataset. Bolded values indicate the highest and those within one standard deviation of the highest in each column.}
        \label{tab:vri}
    \resizebox{\textwidth}{!}{
    \begin{tabular}{cc|ccccc|ccccc}
\toprule
\multirow{2}{*}{Method} & \multirow{2}{*}{} & \multicolumn{5}{c|}{CIFAR-10}                                                                                                                          & \multicolumn{5}{c}{CIFAR-100}                                                                                                                         \\
                           &                   & Sym. 20\% & Sym. 40\% & Sym. 60\% & Asy. 40\% & IDN 40\% & Sym 20\% & Sym 40\% & Sym 60\% & Asy. 40\% & IDN 40\% \\ \midrule
\multirow{2}{*}{VRI-CE}        & best              & \bm{$93.32_{\pm.02}$}              & $91.15_{\pm.05}$              & $87.44_{\pm.24}$              & $91.43_{\pm.16}$               & $90.08_{\pm.30}$                & $71.21_{\pm.03}$              & $65.67_{\pm.47}$              & $57.60_{\pm.13}$               & \bm{$63.28_{\pm.24}$}               & $62.13_{\pm.01}$               \\
                           & ens          & \bm{$93.46_{\pm.59}$}              & \bm{$91.92_{\pm.12}$}              & \bm{$88.27_{\pm.11}$}              & $89.94_{\pm.67}$               & $88.80_{\pm.88}$                & $71.42_{\pm.22}$              & $66.22_{\pm.63}$              & $58.75_{\pm.32}$              & $60.64_{\pm.06}$               & $65.88_{\pm.29}$               \\ \hline
\multirow{2}{*}{VRI-CE(n)}        & best              & $93.49_{\pm.13}$              & $91.44_{\pm.03}$              & $87.94_{\pm.12}$              & $90.38_{\pm.72}$               & $88.71_{\pm.44}$                & $72.08_{\pm.34}$              & $65.69_{\pm.25}$              & $57.15_{\pm.28}$               & $56.88_{\pm.59}$               & $62.96_{\pm.47}$               \\
                           & ens          & $70.86_{\pm.65}$              & $60.14_{\pm1.18}$              & $52.01_{\pm.60}$              & $57.97_{\pm4.71}$               & $60.66_{\pm2.76}$                & $33.50_{\pm1.06}$              & $23.84_{\pm1.01}$              & $15.52_{\pm.72}$              & $24.39_{\pm.18}$               & $24.94_{\pm.74}$               \\ \hline
                           
\multirow{2}{*}{VRI-CE(p)}     & best              & $93.58_{\pm.21}$                     & $91.38_{\pm.08}$                    & $87.52_{\pm.21}$              & $91.77_{\pm.85}$               & $89.43_{\pm.22}$               & $71.87_{\pm.28}$                    & $65.97_{\pm.23}$              & $58.06_{\pm.23}$              & $55.92_{\pm.11}$                     & $62.70_{\pm.02}$                \\
                           & ens          & \bm{$93.95_{\pm.35}$}                    & $90.96_{\pm.18}$                     & \bm{$88.39_{\pm.13}$}              & $89.77_{\pm.20}$                & $87.52_{\pm.94}$               & $72.45_{\pm.01}$                    & $64.03_{\pm.10}$              & $55.47_{\pm2.23}$              & $54.82_{\pm.16}$                     & $58.88_{\pm.81}$               \\ \hline
\multirow{2}{*}{VRI-MAE}       & best              & $93.42_{\pm.02}$              & $91.54_{\pm.08}$              & $87.56_{\pm.12}$              & $91.17_{\pm.27}$               & $88.91_{\pm.38}$               & $71.45_{\pm.34}$              & $65.36_{\pm.25}$              & $57.18_{\pm1.05}$              & $55.36_{\pm1.0}$                & $62.62_{\pm.26}$               \\
                           & ens          & \bm{$93.82_{\pm.02}$}              & \bm{$92.22_{\pm.06}$}              & \bm{$88.39_{\pm.23}$}              & \bm{$91.87_{\pm.45}$}               & $89.56_{\pm.08}$               & \bm{$73.19_{\pm.31}$}              & $67.30_{\pm.11}$              & $58.03_{\pm1.34}$              & $56.70_{\pm1.56}$                & $62.98_{\pm.79}$               \\ \hline
\rowcolor{lightgray}
     & best              & $93.10_{\pm.21}$               & $90.96_{\pm.06}$              & $86.34_{\pm.19}$              & $91.59_{\pm.15}$                     & $90.98_{\pm.25}$                     & $71.85_{\pm.04}$              & $67.13_{\pm.40}$               & $58.85_{\pm.17}$              & $62.42_{\pm1.36}$               & $66.48_{\pm.14}$               \\ \rowcolor{lightgray}
\multirow{-2}{*}{VRI-CLID}              & ens          & \bm{$93.35_{\pm.41}$}              & $90.50_{\pm.04}$               & $86.58_{\pm.31}$              & \bm{$92.10_{\pm.47}$}                     & \bm{$91.10_{\pm.01}$}                      & \bm{$73.40_{\pm.16}$}               & \bm{$68.99_{\pm.11}$}              & \bm{$60.45_{\pm.36}$}              & \bm{$63.70_{\pm1.16}$}                & \bm{$68.20_{\pm.47}$} \\
\bottomrule
    \end{tabular}}

    \end{subtable}


\end{table*}

\subsection{CLID-MU: CLID-based Meta-Update Strategy} 
\label{sec:clid_meta}

This connection between the CLID metric and model performance lays a foundation for using CLID to evaluate models when a guaranteed clean labeled set is not available. We propose to have CLID serve as meta loss in the {\it Meta-Train} step, as illustrated in Figure \ref{fig:method_flow}
The stop-gradient operation is designed primarily to prevent trivial constant solutions. Substituting $L^{clid}$ into Eq. \ref{eq:meta-train}, the 
{\it meta-update step} becomes:

\begin{equation}
    \psi^{t+1} = \psi^t+ \frac{\alpha\gamma}{n}\sum_{i=1}^{n}g_{i}\frac{\partial \Omega(L_{i};\psi^t)}{\partial \psi},
\end{equation}
where $g_{i}=\frac{\partial L_{i}(\theta)}{\partial \theta} \mid_{\theta^{t}}^{T} \frac{\partial L^{clid}(X,\hat{\theta}^{t+1}(\psi))}{\partial \theta} \mid_{\hat{\theta}^{t+1}}$, $L_{i}(\theta)$ denotes cross-entropy loss of $x_i$. $g_i$ represents the similarity between the gradient of the loss for the training sample $x_i$ and the gradient of $L^{clid}$ on the complete unlabeled batch. The meta-model is then updated accordingly. The overall optimization procedure can be found in Algorithm \ref{alg:algorithm} (appendix). We note the potential risk of overfitting to noisy labels during the training process of \modelabbr{}. However, unlike prior meta-update strategies based on supervised loss, our \modelabbr{}  has a reduced risk of overfitting.
\begin{remark}
 Prior approaches aim to minimize the supervised meta loss $L^{meta}$ for $g_{i}=\frac{\partial L_{i}(\theta)}{\partial \theta} \mid_{\theta^{t}}^{T} \frac{\partial L^{meta}(X,\Tilde{Y},\hat{\theta}^{t+1}(\psi))}{\partial \theta} \mid_{\hat{\theta}^{t+1}}$.
 That is, the noisy labels $\Tilde{Y}$ are involved in the meta-model updating step. This can lead to overfitting to these noisy labels. In contrast, our \modelabbr{} does not rely on noisy labels in the meta-update step. As training progresses, the meta-model provides guidance to the classification model, which in turn enhances the meta-model, creating a \textbf{virtuous cycle}.
\end{remark}
\textbf{Computational complexity analysis.} Given a batch of validation data of size $m$, sample-wise supervised evaluation metrics, such as Cross-Entropy (CE) and Mean Absolute Error (MAE), have a computational complexity of $\mathcal{O}(m)$. In contrast, CLID, being a pair-wise metric, has a computational complexity of $\mathcal{O}(m^2)$, making it $m$ times more computationally intensive than supervised metrics. However, we argue that this additional computational cost is practically justified, as CLID-MU obviates the need for a clean validation set and substantially reduces the extensive hyperparameter tuning required by alternative methods (e.g., determining thresholds for selecting pseudo-clean sets using the small-loss criterion).

\subsection{Snapshot Ensembling}
With CLID as an effective evaluation metric for model evaluation, we propose leveraging snapshot ensembling \cite{huang2017snapshot} for inference. Specifically, the top $K$ snapshots (i.e., model weights) are selected based on their CLID scores under CLID-MU, evaluated on the entire meta-dataset, and subsequently saved. During inference, the predictions from all saved snapshots are averaged to generate the final output. Let $y_{i}$ be the one-hot ground truth label of $x_{i}$, and $F_{y_{i}}(x_{i})$ be the predicted probability that $x_{i}$ belongs to $y_{i}$. 
The final prediction, $F(x_{i})$, is the average of the $K$ saved model snapshots $\{f^{k}(x_{i})\}_{k=1}^{K}$. 
The exponential loss $L^{exp}=\frac{1}{n}\sum_{i}^{n}\text{exp}(-F_{y_{i}}(x_{i}))$ is often used to measure the error of the model \cite{wang2020comprehensive}. We analyze the convergence of the snapshot ensembling by presenting the upper bound of the exponential loss.
\begin{theorem} \label{thm:ensemble}
The exponential loss ${L}^{exp}$ is bounded by 
\[
{L}^{exp}\leq\Pi_{k=1}^{K}R_{k}^{1/K}, 
\]
where $R_{k}=\frac{1}{n}\sum_{i=1}^{n}\text{exp}(-f^k_{y_i}(x_{i}))$. The upper bound of $L^{exp}$ decreases as $K$ increases.
\end{theorem}
Theorem \ref{thm:ensemble} demonstrates an exponential decrease in the $L^{exp}$ bound as we save more snapshots.

\section{Experimental Study}
With  CLID-MU being a model-agnostic approach, we conduct experiments to validate its effectiveness on benchmark datasets across various learning methods. We focus on four research questions:
\begin{enumerate}
    \item \textbf{RQ1}: How effective is CLID-MU compared to alternative baselines across diverse meta-learning methods when a clean labeled dataset is unavailable?
    \item \textbf{RQ2}: How does CLID-MU perform compared to state-of-the-art LNL frameworks? 
    \item \textbf{RQ3}: How robust is CLID-MU in real-world scenarios where only a small portion of the data is noisily labeled, while the majority remains unlabeled, i.e., in semi-supervised settings? 
    \item \textbf{RQ4}: Is CLID-MU sensitive to the selection of its hyperparameters? 
\end{enumerate}

\subsection{Comparison with Meta-learning Methods}
This subsection demonstrates that, without requiring access to a clean labeled dataset, our method achieves superior performance across two meta-learning methods: WNet \cite{Shu2019MetaWeightNetLA} and VRI \cite{sun2024variational}.

\textbf{Experimental setup.} Experiments are run on the {\it CIFAR-10} and {\it CIFAR-100} \cite{krizhevsky2009learning} with three types of noise: symmetric, asymmetric, and instance-dependent noise. Symmetric noise uniformly flips labels to a random class with probability $p$. Asymmetric noise means the labels are flipped to similar classes with probability $p$. The instance-dependent noise is obtained by setting a random noise probability $p$ for each instance following a truncated Gaussian distribution \cite{xia2020part}. For all methods, the meta-dataset size is fixed at 1000 samples. CLID-MU uses a randomly sampled meta-dataset from the noisy training set, while baseline methods select it evenly across classes based on training labels. We report the \textit{best} accuracy, defined as the highest accuracy achieved on the clean test set during training. For fair comparisons, we apply snapshot ensembling to all methods and report the \textit{ensemble} accuracy, obtained using five model snapshots selected based on the meta-objective of each method (i.e., evaluation performance).

\textbf{Baselines.}  We take the standard meta-learning with clean meta-data using cross-entropy (CE) loss as the meta-objective to reference ceiling performance. We compare CLID-MU against three alternative baseline methods: 1) \textbf{CE(n)}: Noisy samples are randomly selected with class balance to form the meta-dataset, using CE as the meta-objective. 2) \textbf{CE(p)}: Following \cite{sun2024variational}, we select reliable samples with smaller losses using Gaussian Mixture Model (GMM) clustering, ensuring an even selection across all classes. These samples are designated as the pseudo-clean meta-dataset with CE as the meta-objective. Following prior work, we perform an initial warming-up phase (10 epochs for CIFAR-10 and 30 epochs for CIFAR-100) before proceeding with sample selection. 3) \textbf{MAE}: Following \cite{ghosh2021we}, we set MAE as the meta-objective and apply it to a randomly selected meta-dataset drawn from the noisy training set. See the appendix for implementation details.

\textbf{Results.} Tables \ref{tab:wnet} and  \ref{tab:vri} present the results on CIFAR datasets for the meta-learning methods WNet and VRI, respectively. Our findings demonstrate that CLID-MU is effective when integrated into different meta-learning frameworks. For both methods, While using cross-entropy loss on a noisy validation set yields strong \textit{best accuracy} under simple noise, it leads to substantial degradation in \textit{ensemble accuracy} and overall performance in complex settings. When incorporating other meta-objectives into WNet, we observe that the performance of all methods remains relatively close on CIFAR-10 across different noise settings. WNet-CLID achieves superior performance in high-noise scenarios, including 60\% symmetric noise, 40\% asymmetric noise, and 40\% instance-dependent noise. Impressively, WNet-CLID even outperforms WNet-CE, which leverages a clean meta-dataset, confirming its effectiveness in handling complex and challenging noise patterns. 


In the VRI framework, VRI-CLID demonstrates competitive performance against other baselines under both symmetric and asymmetric noise on CIFAR-10 and excels under instance-dependent noise. On CIFAR-100, VRI-CLID excels across all noise settings, consistently outperforming every baseline. It even surpasses the performance ceiling of VRI-CE, which is trained on a clean meta-dataset. The {\it ensemble} accuracy generally exceeds the {\it best} accuracy in most scenarios, underscoring the benefits of snapshot ensembling during inference. However, in more challenging settings (asymmetric and instance-dependent noise) on CIFAR-10 and CIFAR-100, the ensemble accuracy of VRI-CE(p) falls below its best accuracy, indicating that the pseudo-clean set selected using the small-loss trick may be unreliable for model evaluation.

\subsection{Comparison with State-of-the-art Methods}
We compare our method with competitive methods on datasets with real-world noise, CIFAR-10N (Worst) and CIFAR-100N \cite{wei2021learning} in Table \ref{tab:cifarN}. The meta-dataset with 1000 samples is randomly sampled from the noisy training set. We compare with the competitive methods:(1) Co-teaching \cite{han2018co} and ELR+ \cite{liu2020early} train two networks that mutually refine each other; (2) SOP \cite{liu2022robust} models label noise through over-parameterization and incorporates an implicit regularization term; (3) DivideMix \cite{li2020dividemix} employs two networks, dynamically separating the training set into clean and noisy subsets using the small-loss trick and handling the noisy subset through a semi-supervised learning fashion. To compare with the state-of-the-art methods, we integrated VRI-CLID into DivideMix to demonstrate that our method is compatible with existing LNL methods and enhances performance on both datasets. 

We also experiment on the Animal-10N \cite{song2019selfie} and Clothing1M \cite{xiao2015learning} data sets, both of which contain naturally occurring noisy labels introduced by human error. As shown in Tables \ref{tab:animal} and \ref{tab:clothing1m}, we compare the performance of VRI-CLID and WNet-CLID with state-of-the-art methods. On Animal-10N, VRI-CLID and VRI-CE(p) achieve similar performance to each other, and outperform all competing methods except DISC, demonstrating the effectiveness of CLID-MU in handling real-world label noise. On Clothing1M, both WNet-CLID and VRI-CLID achieve performance comparable to other leading methods and surpass the baseline of using MAE loss as the meta-objective. 


\begin{table}[t!]
\centering
\caption{Test Accuracy (mean and std dev over 3 runs) on CIFAR-10N Worst and CIFAR-100N. * denotes our implementation, other results are from \cite{wei2021learning}. "$\dagger$" means the reported accuracy is from snapshot ensembling.}
\begin{tabular}{lcc}
\toprule
Method                & CIFAR-10N(Worst) & CIFAR-100N                       \\ \midrule
CE                    & 77.69 ± 1.55    & 55.5 ± 0.66                      \\
Co-teaching           & 83.83 ± 0.13    & 60.37 ± 0.27                     \\
SOP                   & {\bf 93.24 ± 0.21}    & 67.81 ± 0.23                     \\
ELR+                  & 91.09 ± 1.60    & 66.72 ± 0.07                     \\
Divide-Mix*            & 90.43 ± 0.57    & 67.04 ± 0.59                     \\ \midrule
$\text{VRI-CLID}^{\dagger}$              & 89.07 ± 0.18       & 67.53 ± 0.34                     \\
$\text{VRI-CLID + Divide-Mix}^{\dagger} $& 90.70 ± 0.11              & {\bf 70.05 ± 0.20} \\ 
\bottomrule
\end{tabular}

\label{tab:cifarN}
\end{table}


\begin{table}[t!]
\centering
\caption{Test Accuracy (mean and std dev over 3 runs) on Animal-10N. Results are directly from the original papers. "$\dagger$" means the reported accuracy is from snapshot ensembling.}
\resizebox{\columnwidth}{!}{
\begin{tabular}{l|c|l|c}
\toprule
Method & Accuracy    & Method             & Accuracy    \\\midrule
CE \cite{englesson2021generalized}     & 79.4 ± 0.14 & GJS \cite{englesson2021generalized}               & 84.2 ± 0.07 \\
GCE \cite{zhang2018generalized}    & 81.5 ± 0.08 & DISC \cite{Li2023DISCLF}               & \textbf{87.1 ± 0.15} \\
SELIE \cite{song2019selfie}  & 81.8 ± 0.09 & Nested Co-teaching \cite{chen2021boosting} & 84.1 ± 0.10 \\
MixUp \cite{zhang2017mixup}  & 82.7 ± 0.03 & $\text{VRI-CE(p)}^{\dagger}$          & 85.5 ± 0.51 \\
PLC \cite{zhang2021learning}   & 83.4 ± 0.43 & $\text{VRI-CLID}^{\dagger}$  (ours)        & 85.6 ± 0.57 \\ \bottomrule

\end{tabular}}
\label{tab:animal}

\end{table}

\begin{table}[t!]
\centering
\caption{Test Accuracy on Clothing1M. Results are directly from the original papers. "$\dagger$" means the reported accuracy is from snapshot ensembling.}
\resizebox{\columnwidth}{!}{
\begin{tabular}{l|c|l|c}
\toprule
Method & Accuracy    & Method             & Accuracy    \\\midrule
CE \cite{yong2022holistic}     & 68.94 & Forward \cite{patrini2017making}              & 69.91 \\
Co-teaching \cite{han2018co}   & 60.15 &  ELR \cite{yuan2025early} & 72.87\\
Dual T \cite{yao2020dual}  & 71.49 & $\text{WNet-MAE}^{\dagger}$   & 72.85\\
BARE \cite{patel2023adaptive}  & 72.28 &  $\text{WNet-CLID}^{\dagger}$  (ours)& \textbf{72.93}\\
VolminNet \cite{li2021provably} & 72.42  &  $\text{VRI-MAE}^{\dagger}$ & 67.78 \\
ROBOT (RCE) \cite{yong2022holistic} & 72.70 &  $\text{VRI-CLID}^{\dagger}$ (ours)& 72.83\\ \bottomrule 
\end{tabular}}
\label{tab:clothing1m} 

\end{table}
\begin{table*}[t!]
\centering
\fontsize{9pt}{9pt}\selectfont
\caption{Test accuracy (mean and std dev over 3 data folds) on CIFAR-10 with different noise types and noise ratios. Average noise ratios of human annotations over these three folds are in parentheses.  Best results in \textbf{bold}, second highest \underline{underlined}.}
\begin{tabular}{lcccccc}

\toprule
\multirow{2}{*}{Methods}  & \multicolumn{2}{c}{Symmetric}                & Asymmetric            & \multicolumn{3}{c}{Human}                                             \\\cmidrule(r){2-3} \cmidrule(lr){4-4} \cmidrule(l){5-7}
          & 20\%                 & 50\%                  & 40\%                  & Aggregate (8.8\%)            & Random1 (16.9\%)            & Worst (40.6\%)               \\ \midrule
UDA                         & 72.04 ± 0.18         & 49.97 ± 2.91          & 70.95 ± 0.27          & 79.97 ± 0.37          & 75.05 ± 0.40          & 63.39 ± 0.89          \\
w/ELR                       & 78.22 ± 0.95         & \underline{63.61 ± 0.33}          & 72.38 ± 0.41          & 80.96 ± 0.28          & 79.14 ± 0.06          & 64.86 ± 3.73          \\
w/MixUp                     & 77.00 ± 0.53         & 58.23 ± 1.49          & \underline{73.27 ± 0.75}    & 82.78 ± 0.31          & 79.52 ± 0.33          & 68.26 ± 0.12          \\
w/WNet-MAE                   & \underline{85.25 ± 0.92}         & 60.79 ± 16.55         & 72.92 ± 0.85          & \underline{86.49 ± 0.84}    & \underline{82.90 ± 0.88}    & \underline{70.72 ± 2.03}    \\
w/WNet-CLID & \textbf{86.22± 1.90} & \textbf{78.73± 3.52}  & \textbf{73.95± 0.32}  & \textbf{88.17 ± 0.81} & \textbf{85.27 ± 1.65} & \textbf{75.09 ± 4.88} \\ \midrule
FixMatch                    & 73.36 ± 0.26         & 51.07 ± 1.10          & 71.76 ± 0.79          & 83.00 ± 0.37          & 78.08 ± 0.54          & 61.37 ± 0.40          \\
w/ELR                       & 74.17 ± 1.56         & 51.00 ± 1.34          & 72.80 ± 0.51          & 83.01 ± 0.48          & 81.07 ± 0.40          & \underline{ 70.10 ± 4.52}    \\
w/MixUp                     & 76.27 ± 0.17         & \underline{58.48 ± 1.32}    & 71.93 ± 0.91          & 83.02 ± 0.21          & 79.27 ± 0.66          & 67.66 ± 0.91          \\
w/WNet-MAE                   & \underline{84.97 ± 3.02}         & 53.43 ± 4.76          & \bf{72.85 ± 0.88} & \underline{86.85 ± 0.18}    & \underline{82.05 ± 0.56}    & 63.32 ± 2.68          \\
w/WNet-CLID & \textbf{88.87± 0.22} & \textbf{84.00± 2.34}  & \underline{72.57± 0.58}     & \textbf{89.78 ± 0.19} & \textbf{88.20 ± 0.86} & \textbf{80.95 ± 2.52} \\ \midrule
FlexMatch                   & 78.49 ± 0.30         & 68.86 ± 1.15          & 76.06 ± 0.42          & 85.01 ± 0.23          & 81.44 ± 0.68          & 72.37 ± 0.86          \\
w/ELR                       & 81.46 ± 0.43         & 63.50 ± 2.08          & 75.78 ± 0.61          & 83.24 ± 0.36          & 81.74 ± 0.87          & 67.20 ± 0.60          \\
w/MixUp                     & 84.73 ± 0.21         & \underline{77.31 ± 0.81}    & 77.84 ± 0.56          & 87.68 ± 0.04          & \underline{85.76 ± 0.60}    & \underline{78.20 ± 0.28}    \\
w/WNet-MAE                   & \underline{87.40 ± 1.23}         & 76.95 ± 6.29          & 78.51 ± 0.25          & \underline{88.52 ± 0.32}    & 85.28 ± 0.63          & 76.79 ± 0.90          \\
w/WNet-CLID  & \textbf{89.29± 0.79} & \textbf{83.57 ± 1.99} & \textbf{78.80± 1.04}  & \textbf{89.71 ± 0.48} & \textbf{88.27 ± 1.15} & \textbf{81.58 ± 2.78} \\
\bottomrule
\end{tabular}

\label{tab:semi_cifar10}
\end{table*}
\subsection{Semi-supervised Real-world Scenarios}
\textbf{Experimental setup.} We conducted experiments on CIFAR-10 with symmetric noise at \{20\%, 50\%\} and asymmetric noise at 40\%, generated following the scheme in \cite{patrini2017making}.
Experiments are also conducted on the real-world human-annotated dataset CIFAR-10N \cite{wei2021learning}. Each image in CIFAR-10N is associated with three kinds of labels:  aggregation of three annotations by majority voting (Aggregate), random selection of one from all annotations (Random 1, 2, 3), and the worst annotation (Worst). The quality of these labels decreases in the mentioned order. We used Aggregate, Random1, and Worst in our experiments. 

\textbf{Baselines.} We evaluated \modelabbr{} by integrating it into three widely-used SSL methods: UDA \cite{Xie2019UnsupervisedTraining}, FixMatch \cite{sohn2020fixmatch}, and Flexmatch \cite{zhang2021flexmatch}. 
We compare it with the following methods,  each also naturally integrated into these SSL frameworks: (1) implicit regularization methods proven to have strong performance in dealing with noisy labels, including ELR \cite{liu2020early} and MixUp \cite{zhang2017mixup}; and (2) WNet-MAE \cite{ghosh2021we}, an explicit regularization method that can 
operate in scenarios without access to clean labeled data.

\textbf{Results.} Table \ref{tab:semi_cifar10} shows the results on CIFAR-10 with various noise types. It can be observed that
WNet-CLID {\it outperforms the compared methods by a large margin across all three SSL methods}, particularly under high noise ratios. ELR and MixUp are less effective under challenging settings, such as symmetric noise at 50\% and with (noisy) human labels. The performance of WNet-MAE degrades with higher noise ratios because it relies on MAE loss, which is heavily dependent on the quality of the labels. In contrast, our WNet-CLID succeeds in eliminating the influence of noisy labels when training the meta-model. The superior performance and robustness under real-world noise demonstrate that it has greater potential to be applied in practical SSL scenarios.

\subsection{Sensitivity Analysis}

\begin{figure}[htbp]
    \centering
    \includegraphics[width=0.7\columnwidth]{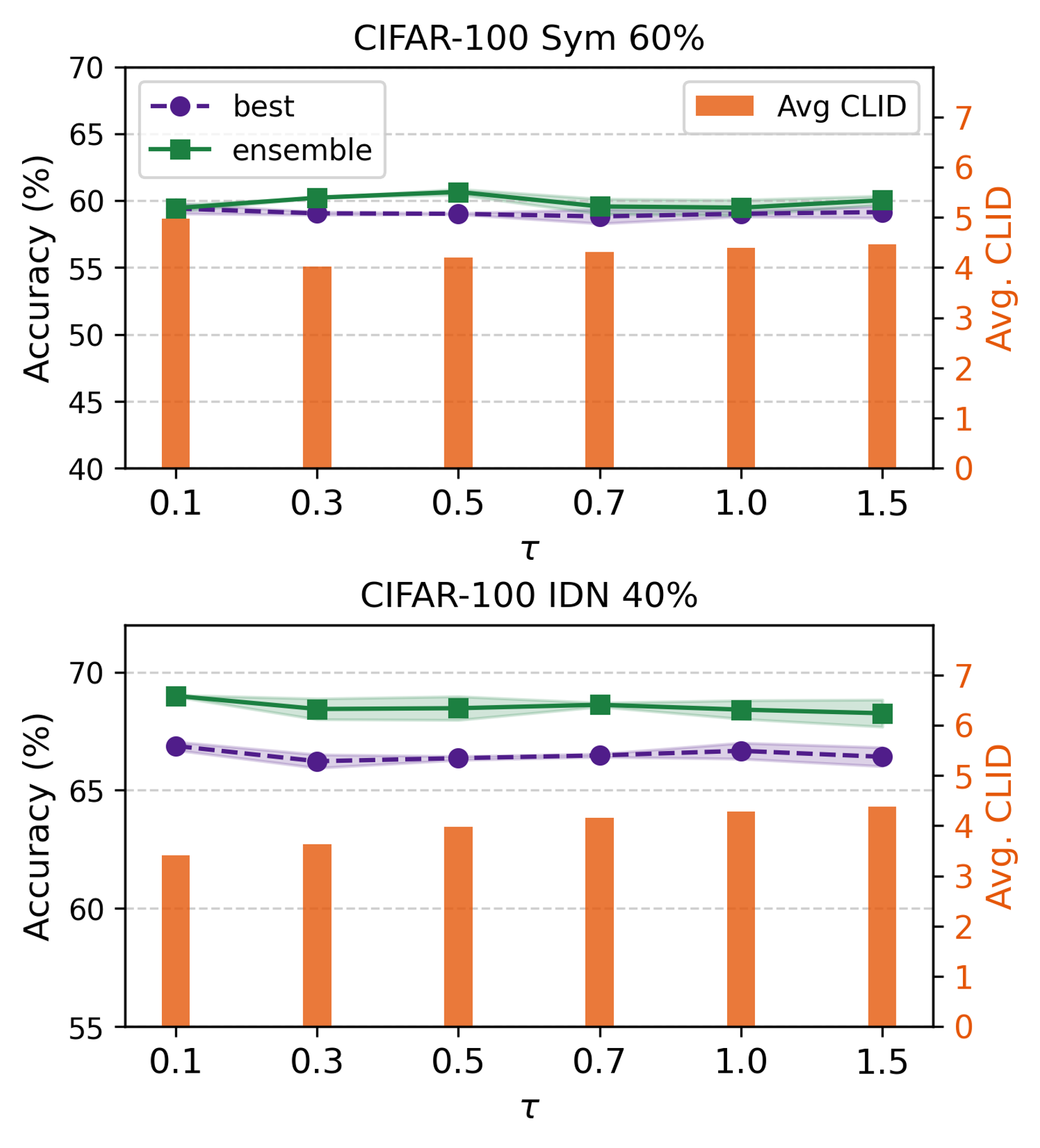}
    \Description{Robustness analysis of CLID-MU across different temperature scaling values ($\tau$) within {0.1, 0.3, 0.5, 0.7, 1, 1.5}, using the VRI meta-learning framework. Experiments are conducted on CIFAR-100 under 60\% symmetric noise and 40\% instance-dependent noise. Both best and ensemble accuracies remain stable across $\tau$ values. The ensemble accuracy peaks at $\tau=0.5$ and $\tau=0.3$ for symmetric noise, and at $\tau=0.1$ for instance-dependent noise. These settings also yield the lowest average CLID scores among the top-5 model snapshots, indicating that $\tau$ can be tuned using CLID to optimize performance.}
    \caption{Test accuracy across different values of temperature scaling ($\tau$) and the corresponding average CLID score of the top-5 model snapshots.}
\label{fig:tau}
\end{figure}
{\bf Effect of temperature scaling.} One critical hyperparameter in CLID is the temperature scaling factor $\tau$. This parameter governs the sharpness of the similarity scores. Since feature embeddings are derived after a ReLU layer, their values are constrained to the range $[0,1]$. When $\tau>1$, the embedding graph becomes more uniform, whereas a smaller $\tau$ amplifies the similarity scores, resulting in a sharper distribution. We evaluate the robustness of CLID-MU across $\tau$ values within \{0.1, 0.3, 0.5, 0.7, 1, 1.5\} using the meta-learning method VRI. The experiments are conducted on the CIFAR-100 dataset with 60\% symmetric noise and 40\% instance-dependent noise. Figure \ref{fig:tau} shows that both the best accuracy and ensemble accuracy are relatively stable across different values of temperature scaling. The ensemble accuracy reaches its highest at $\tau=0.5$ and $\tau=0.3$ for 60\% symmetric noise and $\tau=0.1$ for 40\% instance-dependent noise. Notably, the corresponding average CLID scores of the top-5 model snapshots are the lowest, suggesting that the hyperparameter $\tau$ can be tuned based on average CLID scores to optimize performance.  

\begin{table}[hbt!]
\centering
\fontsize{9pt}{9pt}\selectfont
\caption{Test accuracy with various Meta-Train batch sizes in semi-supervised learning  (Flexmatch) on CIFAR-10N. Results are mean and std dev. across 3 data folds. 
}
\begin{tabular}{lccc}
\toprule
Batch   size & Aggregate   & Random1   & Worst       \\ \midrule
50           & 89.85 ± 0.22 & 88.39 ± 1.12 & 80.28 ± 3.52 \\ \midrule
100          & 89.71 ± 0.48 & 88.27 ± 1.15 & 81.58 ± 2.78 \\ \midrule
300          & 89.83 ± 0.48 & 88.51 ± 1.01 & 81.60 ± 3.33 \\ \midrule
500          & 89.88 ± 0.44 & 88.46 ± 1.02 & 81.67 ± 2.19 \\
\bottomrule
\end{tabular}

\label{tab:aba_batch}
\end{table}

{\bf Effect of batch size for CLID.} We investigated whether a larger batch size for the unsupervised CLID metric could improve training robustness. Using the CIFAR-10N dataset with 4000 labeled data points, we tested FlexMatch with WNet-CLID by varying the meta-train batch size from 50 to 500. As shown in Table \ref{tab:aba_batch}, performance remained stable for batch sizes under low noise ratios (Aggregate and Random1). However, in more challenging settings (Worst), performance improved when the batch size was larger than 50. We observed no significant performance difference for batch sizes of 100 or greater.


\section{Conclusion}
In this paper, we propose \modelfull{} (\modelabbr{}) for learning with noisy labels (LNL) without access to a clean labeled set. Unlike prior works that use supervised loss as meta-loss to evaluate model performance, \modelabbr{} effectively utilizes unlabeled data to measure the cross-layer information divergence (CLID) and then leverages CLID to evaluate the model performance during the Meta-Train step. We evaluate our
\modelabbr{} method on benchmark datasets under synthetic and real-world noises across numerous data settings, including learning with noisy labels and semi-supervised learning with noisy labels. Our comprehensive experimental results demonstrate that our CLID-MU  achieves superior performance compared to state-of-the-art methods. Further, \modelabbr{} is orthogonal to other LNL approaches, such as MixUp and label correction, and can be readily combined with them to enhance their performance. Future work involves exploring CLID for different layers beyond the label space and the feature space of the last encoder block.


\begin{acks}
This work was supported by USDA NIFA (AFRI Award No. 2020-67021-32459) and NSF (Grants IIS-1910880, CSSI-2103832, NRT-HDR-2021871).
\end{acks}

\bibliographystyle{ACM-Reference-Format}

\bibliography{main}

\appendix

\section{Proof for Theorem 1}
\renewcommand{\thetheorem}{1}
\begin{theorem}\label{thm:appendix_ensemble}
The exponential loss ${L}^{exp}$ is bounded by 
\[
{L}^{exp}\leq\Pi_{k=1}^{K}R_{k}^{1/K},
\]
where $R_{k}=\frac{1}{n}\sum_{i=1}^{n}\text{exp}(-f^k_{y_i}(x_{i}))$. The upper bound of $L^{exp}$ decreases as $K$ increases.
\end{theorem}
\begin{proof}
\begin{equation}
\begin{aligned}
    L^{exp} &= \frac{1}{n}\sum_{i=1}^{n}\text{exp}\left(-\frac{1}{K}\sum_{k=1}^{K}f^{k}_{y_{i}}(x_{i})\right) \\
    &=\frac{1}{n}\sum_{i=1}^{n}\prod_{k=1}^{K}\text{exp}\left(-\frac{1}{K}f^{k}_{y_{i}}(x_{i})\right) \\
\end{aligned}
\end{equation}
Since $\text{exp}(-\frac{1}{K}f^{k}_{y_{i}}(x_{i}))\leq\text{exp}(-f^{k}_{y_{i}}(x_{i}))^{\frac{1}{K}}$, we have
\begin{equation}
    \prod_{k=1}^{k}\text{exp}\left(-\frac{1}{K}f^{k}_{y_{i}}(x_{i})\right)\leq \prod_{k=1}^{k}\text{exp}\left(-f^{k}_{y_{i}}(x_{i})\right)^{\frac{1}{K}}
\end{equation}
Thus, the exponential loss can be bounded as:
\begin{equation}
\begin{aligned}
    L^{exp}&\leq\prod_{k=1}^{K}\left(\frac{1}{n}\sum_{i=i}^{n}\text{exp}(-f^{k}_{y_{i}}(x_{i}))\right)^{\frac{1}{K}}\\
    &=\prod_{k=1}^{K}R_{k}^{1/K}
\end{aligned}
\end{equation}
Since $\text{exp}(-f^{k}_{y_{i}}(x_{i}))\leq 1$, we have $R_{k}\leq1$, thus the upper bound will decrease as $K$ increases.  
\end{proof}

\clearpage
\section{Pseudocode}

\addvspace{-2em}
\begin{algorithm}[htpb]
\caption{CLID-based Meta-update Strategy}
\label{alg:algorithm}

\begin{algorithmic}[1] 
\REQUIRE Noisy labeled data $\Tilde{D} $, meta dataset $D^{meta}$, Classification model: $f(\cdot;\theta)$, Meta model: $\Omega(\cdot;\psi)$.\\
\ENSURE Labeled data batch size: $n$, Meta data batch size: $m$, maximum iteration: $T$, temperature scaling: $\tau$, learning rate for classification model: $\alpha$, learning rate for meta model: $\gamma$, number of snapshots to retain: $K$ \\
\STATE Initialize: $t=0$, $\mathcal{M} \gets []$. //{\it M is a bounded list of top-K snapshots and the corresponding CLID scores} \\

\WHILE{$t<T$}
\STATE
  $\{(x_{i} ,\Tilde{y}_{i} )\}_{i=1}^n \gets$ BatchSampler$(\Tilde{D} ,n)$\; \\
  
  $X^{meta} \gets$ BatchSampler$(D^{meta},m)$\;\\
 
\STATE $L_{i}(\theta)=l(f(x _{i};\theta);\Tilde{y}_{i}) \quad $ //{\it cross-entropy loss}\\
//{\it Virtual-Train Step:}
\STATE $\hat{\theta}^{t+1}(\psi)=\theta^{t}-\alpha \frac{1}{n}\sum_{i=1}^{n}\Omega(L_{i};\psi^t)\nabla_{\theta} L_{i}(\theta)\mid_{\theta^{t}}$ \\
//{\it CLID-based Meta-Train Step:} \\
\STATE $\psi^{t+1} =\psi^{t} -\gamma \nabla_{\psi} L^{clid}(X^{meta};\hat{\theta}^{t+1}(\psi))\mid_{\psi^t}$ \\
//{\it Actual-Train Step:} \\
\STATE $\theta^{t+1}=\theta^{t}-\alpha \frac{1}{n}\sum_{i=1}^{n}\Omega(L_{i};\psi^{t+1})\nabla_{\theta} L_{i}(\theta)\mid_{\theta^{t}}$
\\
\IF{EpochEnd(t)}
\STATE $c^{t}=L^{clid}(D^{meta},\theta^{t+1})$ //{\it Evaluate the snapshot}
\IF{$|\mathcal{M}| < K$}
\STATE $\mathcal{M}\gets (\theta^{t}, c^{t}) $
\ELSE
\STATE $(\theta_{\max}, c_{\max}) \gets \arg\max\limits_{(\theta, c) \in \mathcal{M}} c$
\IF{$c^{t} < c_{\max}$}
\STATE $\mathcal{M}\gets (\theta^{t}, c^{t}) $
\ENDIF
\ENDIF

\ENDIF
\STATE $t=t+1$ 

\ENDWHILE
\RETURN $\mathcal{M}$
\end{algorithmic}
\end{algorithm}

\section{Implementation Details}
{\bf Comparison with meta-learning methods.} We do the implementations following the original work WNet and VRI. For VRI, we use PresNet-18 for all noise settings and train the model for 150 epochs. For WNet, we use WRN-28-10 for all noise settings and train the models for 100 epochs. We employ the Cosine Annealing strategy with a 10-epoch period to adjust the learning rate of the classification network. The initial learning rates are 0.02 for PreResNet-18 and 0.05 for WRN-28-10. For the meta-model, we use a learning rate of 0.01 for VRI and $1e^{-5}$ for WNet. Across all experiments, we set the temperature scaling factor ($\tau$) to 0.5, the meta-dataset size to 1000, and the number of model snapshots $K$ to 5. We use a batch size of 100 for both the training set and the meta-dataset.

{\bf Comparison with SOTA.} All experiments on CIFAR-10N and CIFAR-100N are conducted using ResNet-34, following prior works. For VRI-CLID, we employ the Cosine Annealing strategy with a 10-epoch period to adjust the learning rate of the classification network, starting with an initial learning rate (lr) of 0.02. Across all experiments, we set the temperature scaling factor ($\tau$) to 0.5, the meta-dataset size to 1000, and the number of model snapshots $K$ to 5. The batch size is  100 for both the training set and the meta-dataset.

For experiments integrating DivideMix, the initial learning rate is set to 0.03 and decays to 1/10 of its value at 120 and 180 epochs, with a total training duration of 300 epochs. We use a batch size of 128 for the training set and 100 for the meta-dataset.

For experiments on Animal-10N, we use VGG19 to remain consistent with prior works. The initial learning rate is set to 0.1, and we apply the Cosine Annealing strategy with a 160-epoch period for learning rate adjustment. We use a batch size of 128 for the training set and 100 for the meta-dataset.

For experiments on Clothing1M, we use the pre-trained Resnet-50 model to remain consistent with prior works. The learning rate is fixed at 0.0005, and the learning rate for the meta-model is fixed at 0.01. All the models were trained for 10 epochs.

{\bf Semi-supervised learning experiments.} Following the semi-supervised learning benchmarks \cite{wang2022usb}, we used a WRN-28-2 model \cite{Zagoruyko2016WideRN} for all noise settings. The 50,000 training data is split into 4,000 labeled samples and 46,000 unlabeled samples. For CLID-MU, we sampled 1,000 instances from the unlabeled set to construct the meta-dataset, while for WNet-MAE, the meta-dataset was sampled from the labeled set. The classification model was trained using SGD with a momentum of 0.999, a weight decay of $5e^{-4}$, a fixed learning rate of 0.03, and a batch size of 100. The meta-model is trained with a weight decay of $5e^{-4}$ and a batch size of 100 for the meta-dataset. The hyperparameter $\beta$ and meta model learning rate $\gamma$ for UDA, FixMatch, and FlexMatch are set to {7,1,7} and {$1e^{-5}$,$1e^{-4}$,$1e^{-5}$}, respectively.

\section{Computational Complexity} CLID-MU introduces additional computational overhead compared to baseline methods. Empirically, training with CLID-MU on a single NVIDIA A100 GPU requires approximately 140 seconds per epoch using a PreResNet18 backbone, whereas baseline methods complete an epoch in roughly 14 seconds. To mitigate this overhead, we propose several optimization strategies for future work. (1) Instead of computing all pairwise similarities within a batch, we can first construct a sparse class probability graph by connecting each node only to its top-K most similar nodes. The corresponding sparse embedding graph is then built using those connections. This reduces the computational complexity from $\mathcal{O}(m^2)$ to $\mathcal{O}(Km)$, where m is the batch size and $K\ll m$. To further accelerate this step, Approximate Nearest Neighbor (ANN) techniques can be employed using the library FAISS, reducing the complexity to $\sim \mathcal{O}(mlogm)$. (2) Another optimization is to reduce the frequency of meta-model updates by computing the CLID loss once every N steps instead of at every iteration. (3) Meta-model updates may be terminated once the CLID loss converges, thereby eliminating redundant computations in the later stages of training.



\begin{figure*}[htbp]
  \centering

    \begin{subfigure}[htbp]{\linewidth}
    \centering
    \includegraphics[width=0.7\columnwidth]{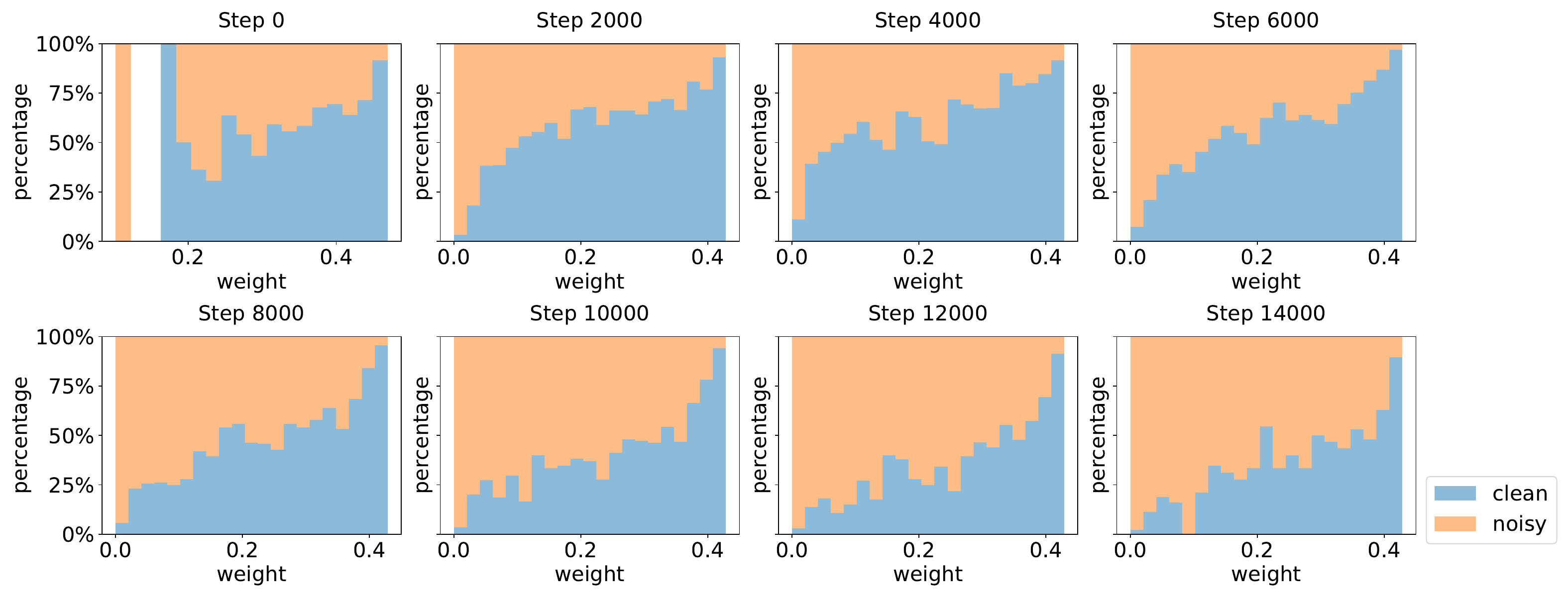}
    \caption{Weight distribution of using CLID for Meta-Train}
    \label{fig:app_wieght_clid}
    \end{subfigure}
  \vspace{0.05cm} 
    \begin{subfigure}[htbp]{\linewidth}
    \centering
    \includegraphics[width=0.7\columnwidth]{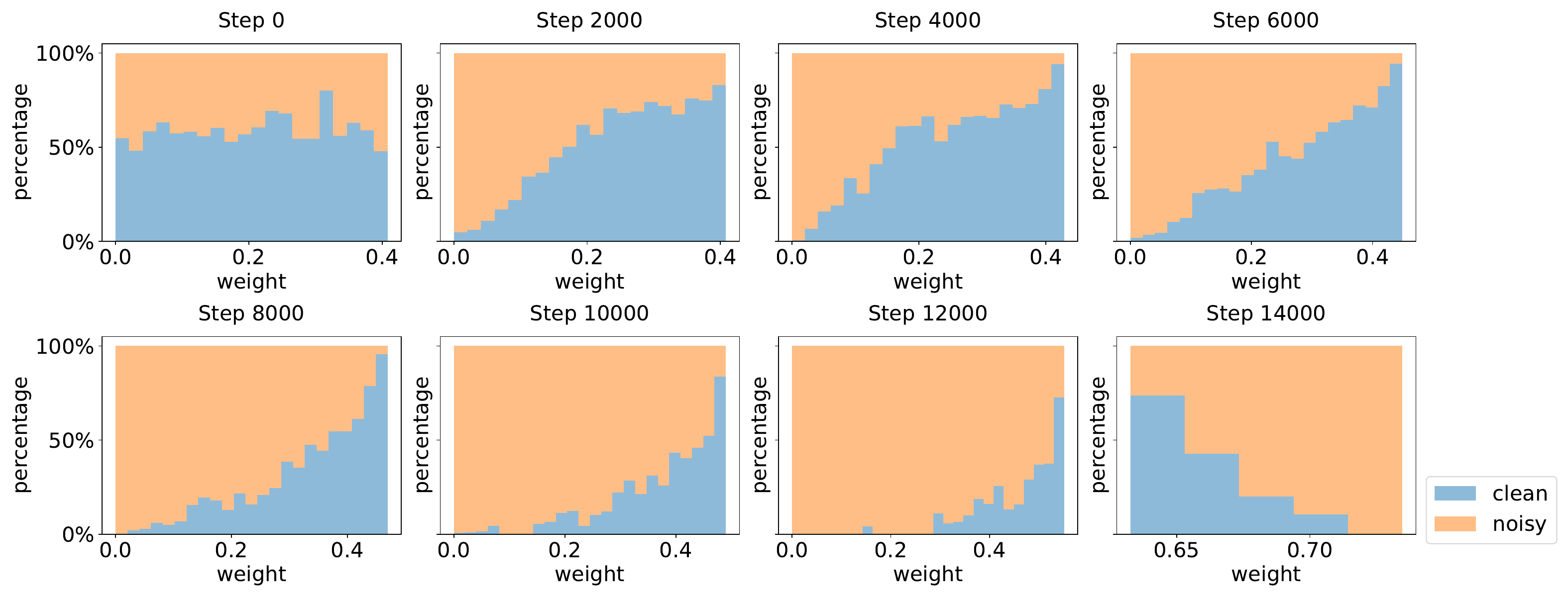}
    \caption{Weight distribution of using MAE for Meta-Train}
    \label{fig:app_wieght_mae}
    \end{subfigure}
  \Description{Weight distribution analysis in the semi-supervised learning setting using FlexMatch. Sample weights are divided into equal-length bins, and the proportion of clean and noisy samples in each bin is visualized. CLID-MU produces more stable weights, with higher weights predominantly assigned to clean samples. In contrast, the MAE-based meta-update increasingly allocates larger weights to noisy samples as training progresses.}
  \caption{The weight distribution of clean and noisy samples in the experiment of FlexMatch on CIFAR-10N (Worst).}
  \label{fig:app_weight}
\end{figure*}

\begin{figure*}[htbp]
  \centering

    \begin{subfigure}[htbp]{\columnwidth}
    \centering
    \includegraphics[width=0.7\linewidth]{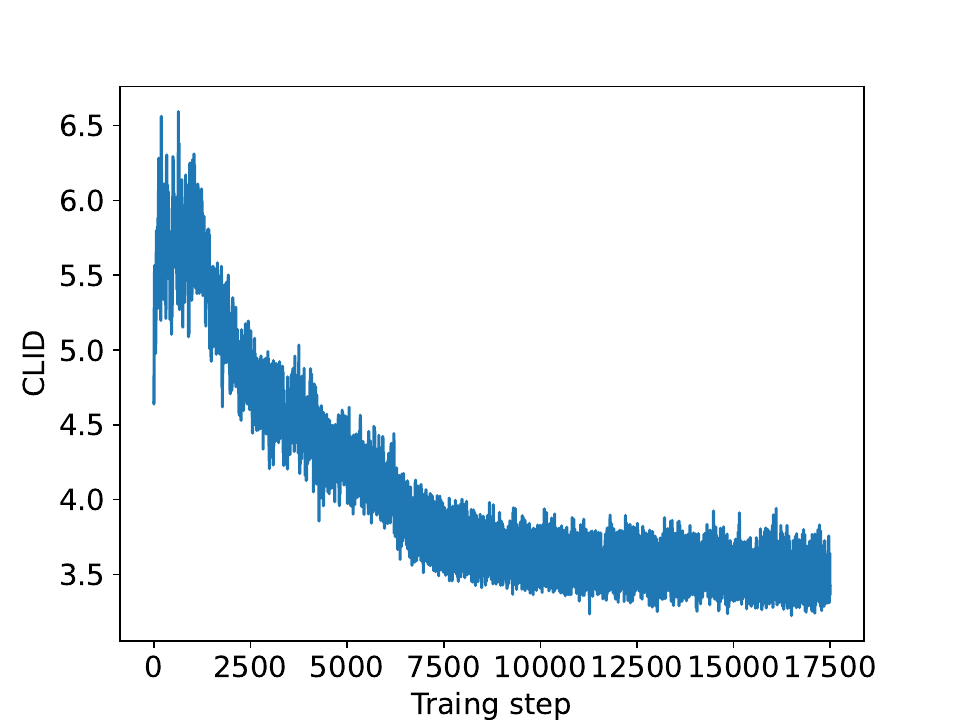}
    \end{subfigure}
  \hspace{0.01cm} 
    \begin{subfigure}[htbp]{\columnwidth}
    \centering
    \includegraphics[width=0.7\linewidth]{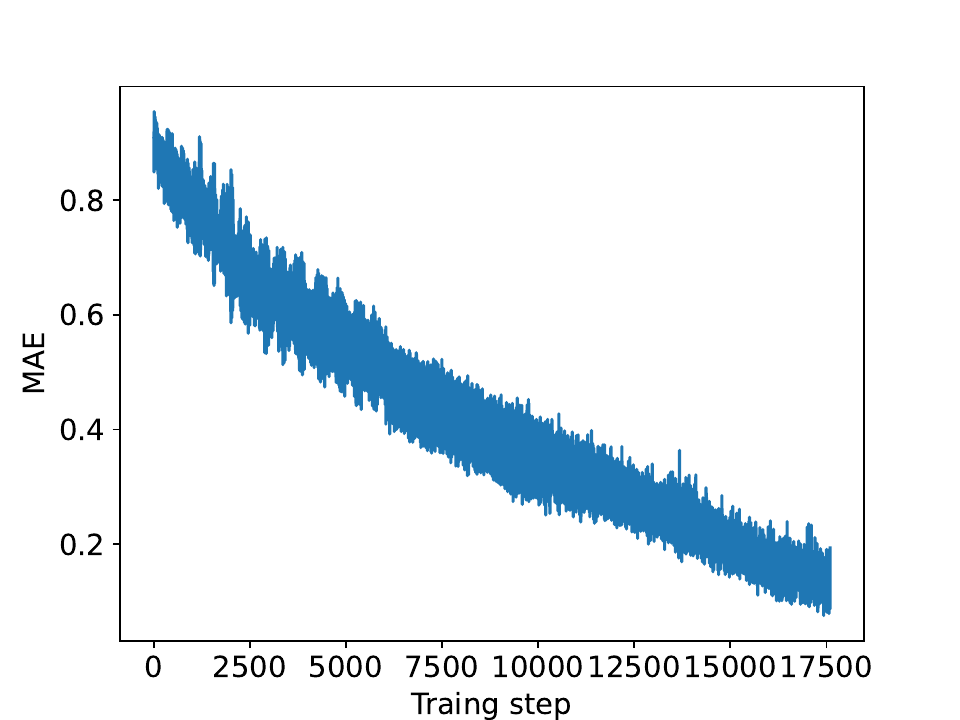}
    \end{subfigure}
  \Description{The trend of meta loss in the Meta-Train step in the experiment of FlexMatch on CIFAR-10N Worst label. Left: our CLID-based Meta-update strategy. Right: MAE-based Meta-update strategy using noisy labeled data as the meta dataset. The MAE loss has not yet converged, indicating that MAE struggles to measure model performance under complex noise patterns effectively; in contrast, CLID converges quickly and remains stable.}
  \caption{Trend of meta loss used in the Meta-Train step in the experiment of FlexMatch on CIFAR-10N Worst label. Left: our CLID-based Meta-update strategy. Right: MAE-based Meta-update strategy using noisy labeled data as the meta dataset.}
  \label{fig:app_loss}
\end{figure*}

\section{Weight Distribution}
 Using the semi-supervised learning experiment with FlexMatch as an example, we divided the weights into equal-length bins and visualized the percentage distribution of clean and noisy samples in each bin, as shown in Figure \ref{fig:app_weight}. The weights generated by CLID-MU are more stable, with most of the larger weights being assigned to clean samples. In contrast, the weights generated by the MAE-based Meta-update tend to increasingly assign higher weights to noisy samples as training progresses. \vfill\eject This phenomenon is not due to the larger learning rate used in training with MAE, nor is it a result of overfitting. By examining the trend of the meta loss in Figure \ref{fig:app_loss}, we observed that the MAE loss has not yet converged, indicating that MAE struggles to effectively measure model performance under complex noise patterns.

\end{document}